\documentclass[5p,twocolumn]{elsarticle}

\usepackage{hyperref}
\newtheorem{theorem}{Theorem}
\usepackage{amsmath,amssymb,amsfonts,multirow,wrapfig,color,bbding,subfigure,diagbox}
\usepackage{booktabs}
\usepackage{ulem}
\usepackage{threeparttable}
\usepackage{graphicx}
\usepackage{textcomp}
\usepackage{makecell}

\begin{document}

\begin{frontmatter}

\title{Weakly Supervised Vessel Segmentation in X-ray Angiograms by Self-Paced Learning from Noisy Labels with Suggestive Annotation}

\author[1,4,5]{Jingyang Zhang}
\author[2]{Guotai Wang}
\author[3]{Hongzhi Xie\corref{mycorrespondingauthor}}
\cortext[mycorrespondingauthor]{Corresponding author}
\ead{xiehongzhi@medmail.com.cn}
\author[3]{Shuyang Zhang}
\author[5]{Ning Huang}
\author[5]{Shaoting Zhang}
\author[1,4]{Lixu Gu\corref{mycorrespondingauthor}}
\ead{gulixu@sjtu.edu.cn}

\address[1]{School of Biomedical Engineering, Shanghai Jiao Tong University, Shanghai, China}

\address[2]{School of Mechanical and Electrical Engineering, University of Electronic Science and Technology of China, Chengdu, China}

\address[3]{Department of Cardiothoracic Surgery, Peking Union Medical College Hospital, Beijing, China}

\address[4]{Institute of Medical Robotics, Shanghai Jiao Tong University, Shanghai, China}

\address[5]{SenseTime Research, Shanghai, China}

\begin{abstract}
The segmentation of coronary arteries in X-ray angiograms by convolutional neural networks (CNNs) is promising yet limited by the requirement of precisely annotating all pixels in a large number of training images, which is extremely labor-intensive especially for complex coronary trees. To alleviate the burden on the annotator, we propose a novel weakly supervised training framework that learns from noisy pseudo labels generated from automatic vessel enhancement, rather than accurate labels obtained by fully manual annotation.
A typical self-paced learning scheme is used to make the training process robust against label noise while challenged by the systematic biases in pseudo labels, thus leading to the decreased performance of CNNs at test time.
To solve this problem, we propose an annotation-refining self-paced learning framework (AR-SPL) to correct the potential errors using suggestive annotation. An elaborate model-vesselness uncertainty estimation is also proposed to enable the minimal annotation cost for suggestive annotation, based on not only the CNNs in training but also the geometric features of coronary arteries derived directly from raw data. Experiments show that our proposed framework achieves 1) comparable accuracy to fully supervised learning, which also significantly outperforms other weakly supervised learning frameworks; 2) largely reduced annotation cost, i.e., 75.18\% of annotation time is saved, and only 3.46\% of image regions are required to be annotated; and 3) an efficient intervention process, leading to superior performance with even fewer manual interactions.
\end{abstract}


\end{frontmatter}

\section{Introduction}
\label{sec:introduction}
Coronary artery disease (CAD) is one of the leading causes of death globally~\cite{vos2016global}. It is primarily caused by obstructive atherosclerotic plaque~\cite{sangiorgi1998arterial}, which narrows the inner wall of coronary artery and decreases normal myocardial perfusion, leading to symptoms such as angina and even myocardial infarction~\cite{reed2017acute}. Percutaneous coronary intervention (PCI) is a minimally invasive surgery to effectively treat CAD in clinical practice. In such a procedure, a cardiologist delivers a catheter with a premounted stent through coronary arteries to the stenosis lesion. Once the lesion is reached, the stent is deployed against the narrow coronary wall by inflating the delivery balloon. Since target vessels are not directly visible, PCI is performed under image guidance by using X-ray angiography to visualize coronary arteries for the injection of radiopaque contrast agent.
The accurate segmentation of vessels in X-ray angiograms (XAs) enables the quantitative analysis of coronary trees~\cite{chen2002quantitative} and is fundamental for the safe navigation of intervention devices for PCI surgery.

\begin{figure}[t]
    \centerline{\includegraphics[width=\columnwidth]{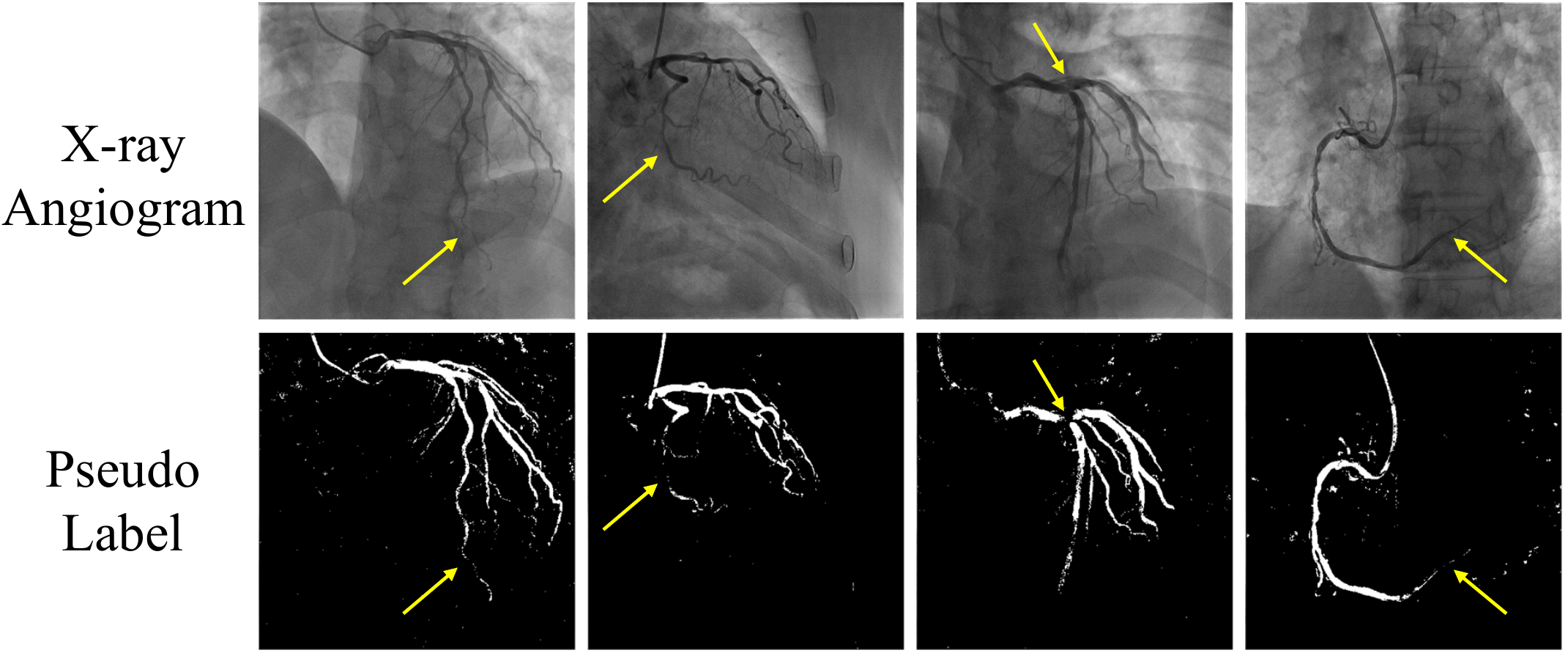}}
    \caption{Noisy pseudo labels generated from vessel enhancement, where the systematic errors are highlighted by yellow arrows.}
    \label{systematic_error}
\end{figure}

Deep learning with convolutional neural networks (CNNs) has achieved the state-of-the-art performance for medical image segmentation~\cite{unet,dcan,nablanet}, including vessel segmentation in XA~\cite{cnnvesselseg1,cnnvesselseg2}.
Following the fully supervised learning framework, its success relies heavily on a large amount of precise annotations for all pixels in training images to improve generalization capability for unseen testing images.
However, precisely annotating coronary arteries is costly and requires special expertise, especially for thin branches with tubular appearance and low contrast in XA.
To alleviate such heavy annotation burden on the annotator, reducing the amount of precise manual annotations is highly demanded in clinical practice~\cite{de2018clinically}.
In contrast, obtaining noisy pseudo labels appears to be less expensive. Specifically, vessel enhancement \cite{vcrpca} automatically extracts vascular structures based on handcraft priors~\cite{OriginalRPCA}, providing a feasible method for generating pseudo labels for training CNNs without any manual interaction. This can largely reduce the manual annotations required for model training, while leading to noise with systematic biases in pseudo labels for structures, such as bifurcation points and thin vessels with small scales, as shown in Fig.~\ref{systematic_error}.
These noisy pseudo labels challenge the learning process and cause performance degradation of CNNs at test time \cite{memorization}. It is desirable to develop a robust training framework against systematic label noise and facilitate segmentation performance close to the fully supervised learning framework.

Aimed at robustly learning from noisy labels, some previous weakly supervised training frameworks model label noise explicitly as an additional network layer~\cite{dgani2018training,simplenl,complexnl} or implicitly using prior knowledge~\cite{bootstrapping,mirikharaji2019learning}. Among them, researchers have shown that the self-paced learning paradigm can be substantially effective and scalable \cite{SPFTN}, owing to its predefined self-paced regularizer \cite{spl}.
This learning paradigm typically assumes a plain distribution of label noise without systematic biases to specific segmentation regions and semantic categories.
An iterative optimization process is used to facilitate noise robustness of the model. In each iteration, the self-paced regularizer progressively selects only easy pixels while excluding difficult pixels with potential label noise from model training. Noisy labels are modified automatically by updating the segmentation results of training images based on the current model. They are expected to contain fewer errors than those in previous iterations, providing improved supervision for the next iteration.
Unfortunately, this self-paced learning paradigm may make the model overfit on easy pixels, leading to a poor generalization performance at test time.
Moreover, the noise in pseudo labels often contains specific biases due to the inherent limitations of vessel enhancement-based generation process. Using this naive self-paced learning paradigm alone has only the limited ability to correct the erroneous pseudo labels.

Manually detecting and correcting potentially erroneous pseudo labels is a practical way to avoid the self-paced learning being corrupted by systematic errors, while it is still labor-intensive and time-consuming.
Suggestive annotation \cite{suggestiveannotation} has been shown to be a more efficient method for interactive refinement by intelligently selecting a small number of the most valuable pixels and then querying their labels. It suggests the annotator accurately label only the most uncertain pixels with potentially incorrect labels \cite{testtimeAug}, commonly based on the widely used model uncertainty \cite{modeluncertainty}, i.e., the entropy of CNNs. The required annotation cost can be successfully reduced owing to the effective exploration of potential errors. However, model uncertainty fails to exploit geometric features derived directly from training images, resulting in redundancy among queries \cite{rep+uc,idnknow} and a low efficiency for manual interaction.
In contrast, considering the vesselness of pixels is expected to lead to more context-aware uncertainty estimation as it takes advantage of vascular geometric features. Since the model uncertainty and vesselness uncertainty are complementary, we believe that their combination would provide more reliable uncertainty estimation that efficiently guides user interaction in suggestive annotation.

To solve these problems, this paper develops a novel weakly supervised vessel segmentation framework, which learns from cost-free but noisy pseudo labels generated from automatic vessel enhancement. Specifically, to overcome noisy pseudo labels with systematic biases, we propose to progressively guide the naive self-paced learning with auxiliary sparse manual annotations, which is called annotation-refining self-paced learning (AR-SPL). AR-SPL not only exploits the available knowledge from noisy pseudo labels, but also corrects potential errors using their corresponding manual annotations. These manual annotations, even when sparse in training images, play an important role in hedging the risk of learning from noisy pseudo labels.
Furthermore, to enable a minimal set of annotations, we propose a model-vesselness uncertainty estimation for suggestive annotation, which dynamically and compactly takes into account the trained CNN and the geometric features of coronary arteries in XAs.

\subsection {Contributions}
The contributions of this work are three-fold.
\begin{itemize}
\item First, we propose a novel weakly supervised learning framework in the context of vessel segmentation, aiming to safely learn from noisy pseudo labels generated by vessel enhancement without performance deterioration at test time.
\item Second, to deal with the biased label noise, we develop online guidance for the naive self-paced learning based on sparse manual annotations, which is crucial for a significant segmentation performance boost.
\item Third, towards minimal manual intervention, we propose a customized vesselness uncertainty based on vascular geometric feature, and then couple it with the widely used model uncertainty by a dynamic tradeoff for more efficient suggestive annotation.
\end{itemize}

Experiments demonstrate the effectiveness and efficiency of the proposed framework, where only a very small set of manual annotations can lead to an accurate segmentation result that is comparable to the fully supervised learning.

\subsection{Related Works}
\subsubsection{Vessel Segmentation in XA}
In the past two decades, a wide range of methods have been proposed to segment coronary arteries in XAs, including the active contour model \cite{activecontour}, level set \cite{levelset} and random walker \cite{randomwalker}. Most of these methods are semi-automatic and sensitive to the initialization of interaction, leading to the lack of robustness and accuracy when faced with nununiform illumination and opaque background structures. Recently, vessel segmentation in XAs has been dominated by deep learning with CNNs, such as a multiscale CNN architecture \cite{cnnvesselseg1} with fully convolutional connections and a multistage framework \cite{cnnvesselseg2} to reduce motion artifacts in the background. However, they all follow the fully supervised learning scheme, which requires precise annotations for all pixels in a large number of training images. To the best of our knowledge, in the context of vessel segmentation, there is no previous perspective that focuses on weakly-supervised learning from noisy labels, i.e., generated by vessel enhancement \cite{vcrpca}.

\subsubsection{Learning from Noisy Labels}
For deep learning with CNNs, noise in training labels inevitably leads to performance degradation at test time \cite{memorization}. How to improve robustness of CNNs when learning from noisy labels is worthy of exploration. This challenge is especially significant yet under-studied for medical image analysis. 
An explicit noise model is constructed in \cite{dgani2018training} to overcome the unreliable noisy annotations for breast lesion, which uses a constrained linear layer \cite{simplenl} and a noise adaptation layer \cite{complexnl}. 
Some other works implicitly treat noisy labels as statistical outliers based on prior knowledge. For example, perceptual consistency proposed in bootstrapping \cite{bootstrapping} is used to augment and modify noisy labels to mitigate their potential degradation. This consistency-based prior knowledge further inspires the label-noise-robust method \cite{min2019two} for cardiac and glioma segmentation in MRI, where model updating is only performed on data samples with inconsistent predictions in the two-stream module. 
Learning difficulty is another prior knowledge to identify noisy samples and then down-weight them during training. Pixel-wise down-weighting strategy shows robustness for highly inaccurate annotations for skin lesion segmentation \cite{mirikharaji2019learning} and thoracic organ segmentation \cite{zhu2019pick}. Towards higher effectiveness and scalability, self-paced learning \cite{spl,SPFTN,prior-knowledge} uses a curriculum setting (also called self-paced regularizer), where learning difficulty is updated in parallel with network parameters via alternating optimization. Despite an elegant theoretical proof, these methods fail to fit the intractable label noise in vessel enhancement, which exhibits more complicated and systematic characteristics.

\subsubsection{Suggestive Annotation with Uncertainty Estimation}
Suggestive annotation \cite{suggestiveannotation} is proposed to choose partial training data for labeling, aimed at a better model performance given a limited annotation budget.
In general, there are two main types: geometry sampling and uncertainty sampling. Geometry sampling queries samples based on the geometric distribution of training data, such as representativeness \cite{huang2010active} among unlabeled samples and diversity \cite{coreset} from labeled samples. However, these distribution measures are challenged by the highly imbalanced foreground and background in XAs.
In addition, uncertainty sampling queries the most uncertain samples for their labels commonly based on model uncertainty \cite{modeluncertainty}, which is also called epistemic uncertainty for CNNs. The accurate estimation of model uncertainty relies on the computationally infeasible Bayesian networks, which can be approximated using Monte Carlo sampling with dropout at test time \cite{modeluncertainty}. Model uncertainty proves a strong relationship with prediction errors \cite{testtimeAug} and thus a promising ability to reduce manual annotations, while it is limited by redundant queries especially during an early training stage \cite{rep+uc,idnknow}. To the best of our knowledge, to overcome this drawback, this paper is the first work to take into account the geometric vascular feature for uncertainty estimation, which acts as an auxiliary cue for the commonly used model uncertainty.

\section{Method}

\begin{figure*}[t]
\centerline{\includegraphics[width=7in]{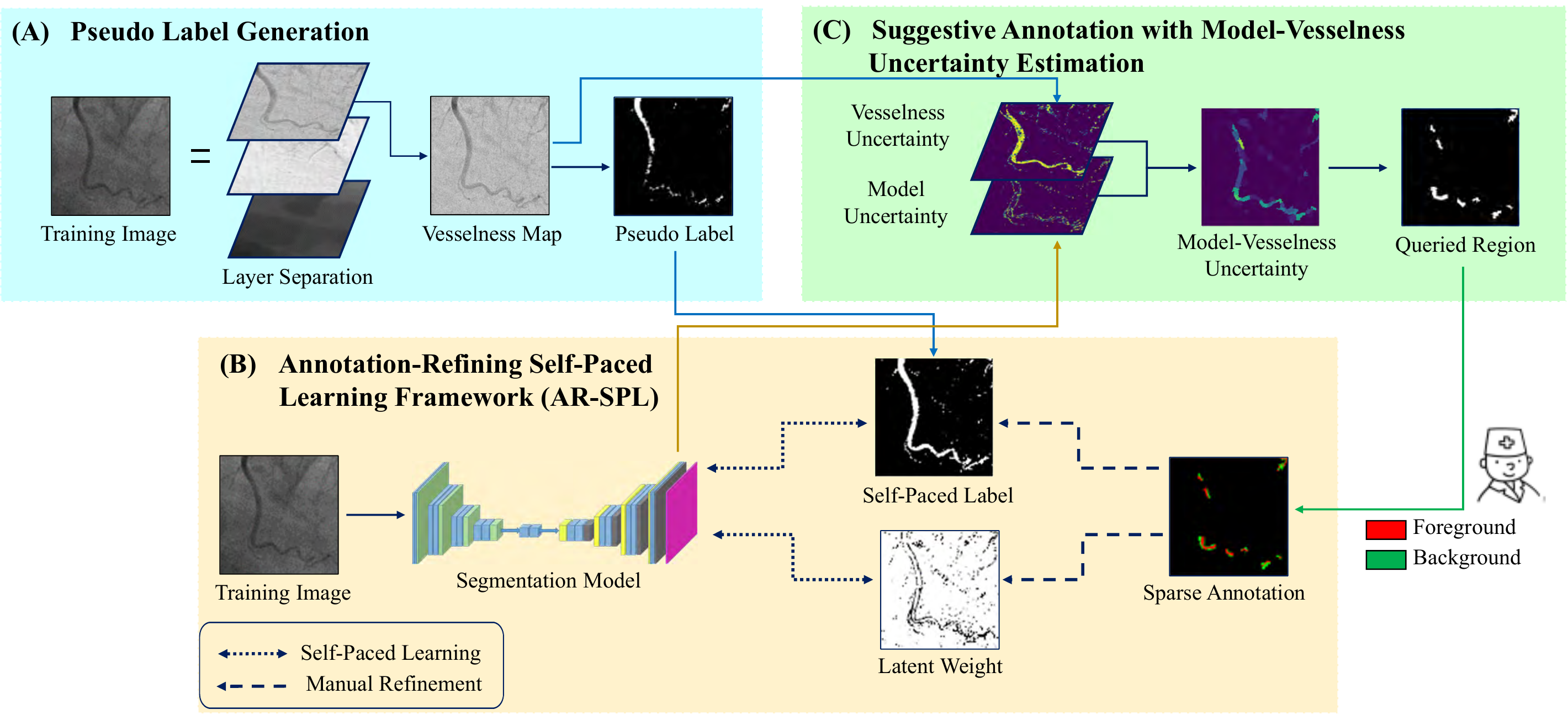}}
\caption{Flow chart of the proposed weakly supervised training framework, which consists of three modules: (A) pseudo label generation; (B) annotation-refining self-paced learning framework (AR-SPL); and (C) suggestive annotation with model-vesselness uncertainty estimation.}
\label{flowChart}
\end{figure*}

The proposed weakly supervised training framework is depicted in Fig.~\ref{flowChart}. It consists of three major parts: (A) pseudo label generation based on automatic vessel enhancement; (B) an annotation-refining self-paced learning framework (AR-SPL) that learns from pseudo labels with online manual refinement based on sparse annotations; and (C) suggestive annotation with model-vesselness uncertainty estimation to enable minimal annotation cost for sparse annotations. Our framework is flexible because it imposes few assumptions on network structure and can be compatible with any popular CNN-based segmentation backbone. Once this training process is completed, a testing process obtains segmentation prediction by performing forward-propagation without human interactions.

\subsection{Pseudo Label Generation} \label{PseudoLabelGeneration}
Although precise labels are fundamental for training a CNN for vessel segmentation, it is highly laborious to obtain them by manually annotating all pixels in a large number of training images.
Vessel enhancement provides a cost-free but noisy alternative for precise labels, called pseudo labels, so as to largely reduce annotation cost as compared with fully manual annotation.
It extracts coronary arteries automatically yet coarsely from background, returning a vesselness map that quantitatively measures vascular structures.

Towards a comprehensive leverage of temporal and appearance priors of coronary arteries, layer separation \cite{vcrpca} is a promising method for vessel enhancement, which separates the original XA into three independent layers, such as a large-scale structure layer, a quasi-static background layer and a vessel layer that contains coronary arteries.
Specifically, we first subtract the large-scale structure layer from the original XA by a morphological closing operation, and obtain a difference image containing the target coronary arteries and residual quasi-static background structures with small scales. Then, robust principle component analysis (RPCA) \cite{OriginalRPCA} is used to further separate the difference image into a quasi-static background layer and a vessel layer based on the quasi-static motion constraint and sparse appearance constraint, respectively. For each training image $\mathbf{x}_{i}$, in order to take advantage of the beneficial temporal cue, layer separation is performed offline on the entire temporal sequence $\mathbf{X}_{i}$, which contains  $\mathbf{x}_{i}$ as a contrast-filled frame\footnote{During PCI, a cardiologist commonly acquires an XA sequence rather than one single XA frame, recording the inflow and fade of contrast agent through coronary arteries.  However, only the key frame with the contrast-filled vessels is used for segmentation in this study.}. The decomposition of the vessel layer sequence  $\mathbf{S}_{i}$ and background layer sequence $\mathbf{L}_{i}$ via RPCA is formulated as follows:
\begin{equation}\label{RPCA_formulation}
    {\min\limits_{\mathbf{L}_{i},\mathbf{S}_{i}~}{\left\| \mathbf{L}_{i} \right\|_{*} + \xi\left\| \mathbf{S}_{i} \right\|_{1}}} \hskip 1pc \text { s.t. }~{\mathbf{D}\mathbf{I}}_{i} = \mathbf{L}_{i} + \mathbf{S}_{i}
\end{equation}
where ${\mathbf{D}\mathbf{I}}_{i}$ is the sequence of difference images acquired by the morphological closing operation on $\mathbf{X}_{i}$. $\left\| \cdot \right\|_{*}$ and $\left\| \cdot \right\|_{1}$ are the nuclear norm and $l_{1}$  norm, respectively. The regularization parameter $\xi$ controls the tradeoff between them, indicating the capability of extracting candidate coronary arteries in the separated vessel layers. Objective function Eq.~\ref{RPCA_formulation} has proven to be convex and can be solved by an inexact augmented Lagrange multiplier method \cite{inexactALM}.

\begin{figure}[t]
\centerline{\includegraphics[width=\columnwidth]{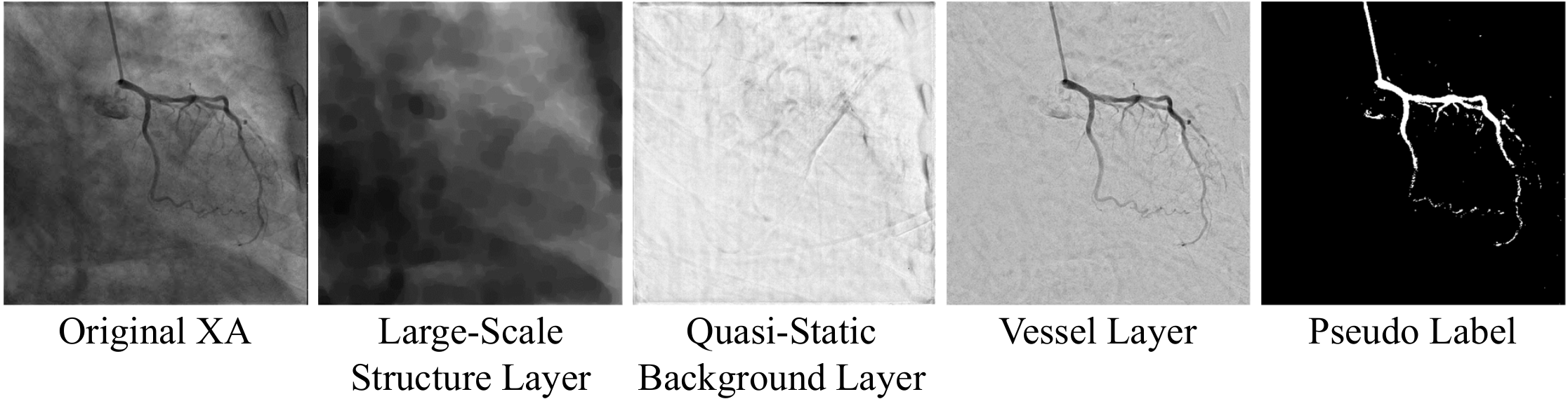}}
\caption{An example of pseudo label generated from layer separation on the original XA.}
\label{LayerSeparation}
\end{figure}

After the RPCA decomposition, the separated vessel layer $\mathbf{s}_{i}$ is treated as the vesselness map for $\mathbf{x}_{i}$ by selecting the corresponding frame from $\mathbf{S}_{i}$. Finally, we apply the Otsu thresholding to $\mathbf{s}_{i}$, generating the pseudo label  $\overline{\mathbf{y}}_{i}$ (an example is shown in Fig.~\ref{LayerSeparation}) that will be used in the following AR-SPL to train a CNN for vessel segmentation.

\subsection{Annotation-Refining Self-Paced Learning (AR-SPL)}
Pseudo labels are obtained automatically by vessel enhancement based on handcrafted priors in layer separation, which leads to inevitable noise due to complex background and inhomogeneous contrast inflow. These systematic rather than random label noise may deteriorate the training of CNN for vessel segmentation if no additional strategies are applied.

\subsubsection{Naive Self-Paced Learning Scheme}

\begin{figure}[t]
\centerline{\includegraphics[width=\columnwidth]{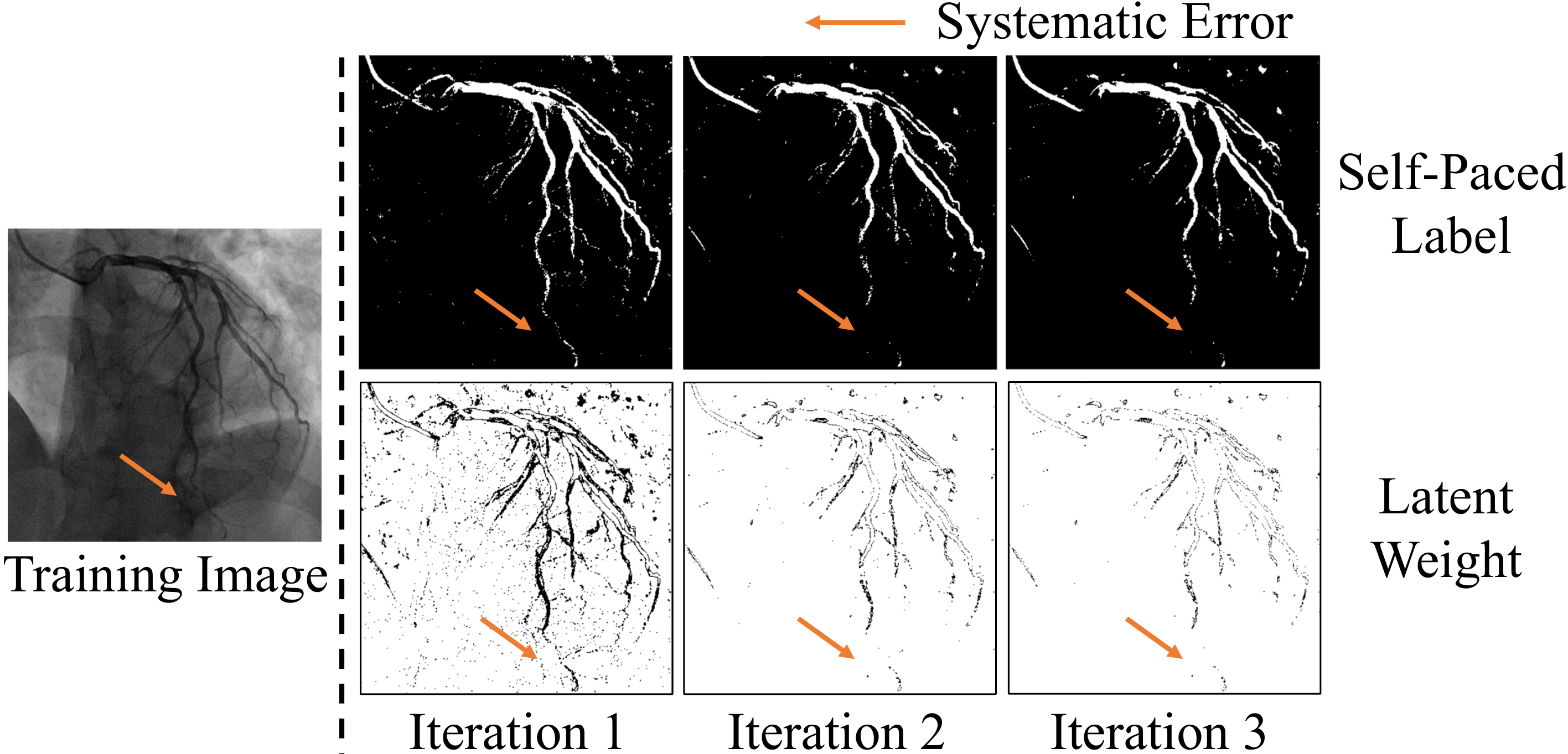}}
\caption{Examples of self-paced labels and latent weights in iterations $1$, $2$ and $3$ of alternating minimization for the naive self-paced learning. Note that the systematic errors are maintained and even amplified, as highlighted by the orange arrows.}
\label{SPLFigure}
\end{figure}
We adopt a self-paced learning scheme \cite{spl} to overcome the negative effect of label noise on model training. It is inspired by the cognitive processes of humans and animals, where a CNN is learned gradually from pixels ranked in descending order of learning difficulty while excluding difficult pixels with potentially noisy labels. This progressive training paradigm enables the model training to focus on easy pixels whose labels have a higher chance to be correct.
Formally, consider the vessel segmentation task in XAs with a training set $\left\{\left(\mathbf{x}_{1}, \overline{\mathbf{y}}_{1}\right), \cdots,\left(\mathbf{x}_{m}, \overline{\mathbf{y}}_{m}\right)\right\}$, where $\mathbf{x}_{i} \in \mathbf{R}^{n}$ denotes training image $\mathbf{x}_{i}$ with $n$ total pixels and ${\overline{\mathbf{y}}_{i} \in \left\{ 0,1 \right\}^{n}}$ denotes its corresponding binary noisy pseudo labels obtained from vessel enhancement. We formulate the self-paced learning scheme as a minimization problem:
\begin{equation}\label{SPL_Obj}
{\begin{array}{c}{\underset{\boldsymbol{V}, \boldsymbol{\boldsymbol{Y}}, \boldsymbol{W}}{\operatorname{argmin}} \sum_{i}^{m} \mathbf{v}_{\boldsymbol{i}} L\left(\mathbf{y}_{i}, \boldsymbol{\Phi}\left(\mathbf{x}_{i} | \boldsymbol{W}\right)\right)+f(\boldsymbol{V} ; {\tau},{\gamma})+\lambda\|\boldsymbol{W}\|_{2}} \\ {\text { s.t. } \boldsymbol{V} \in[0,1]^{n \times m}}\end{array}}
\end{equation}
where $\boldsymbol{W}$ denotes the model parameters of the CNN.
$\boldsymbol{V} = \left\lbrack {\mathbf{v}_{1},\mathbf{v}_{2},\ldots,\mathbf{v}_{m}} \right\rbrack$ represents the latent weights for all $m$ training images, in which $\mathbf{v}_{i} \in \left\lbrack 0,1 \right\rbrack^{n}$ is related to pixel-wise learning difficulty for $\mathbf{x}_{i}$. It is empirically initialized based on the obtained vesselness maps, as described in Section \ref{Implementation}.
Intuitively, the easier a pixel is, the less likely it is to have label noise: a higher latent weight should be assigned in this case. This relationship is formulated as a self-paced regularizer $f\left( {\boldsymbol{V};\tau,\gamma} \right)=-\tau\|\boldsymbol{V}\|_{1}-\gamma\|\boldsymbol{V}\|_{2,1}$ \footnote{It exhibits better performance than other state-of-the-art self-paced regularizers, as shown in Supplementary Materials.}, where the easiness term (the negative $l_1$ norm: $-\|\boldsymbol{V}\|_{1}=-\sum_{i}^{m}\left\|\mathbf{v}_{i}\right\|_{1}$) implicitly models the relationship between learning difficulty and latent weight, and the diversity term \cite{diversity1} (the negative $l_{2,1}$ norm $-\|\boldsymbol{V}\|_{2,1}=-\sum_{i}^{m}\left\|\mathbf{v}_{i}\right\|_{2}$) improves diversity between latent weights for more comprehensive knowledge. $\tau$ and $\gamma$ are hyperparameters imposed on these two terms, which control the learning pace during model training.
In addition, $\boldsymbol{Y} = \left\lbrack {\mathbf{y}_{1},\mathbf{y}_{2},\ldots,\mathbf{y}_{m}} \right\rbrack$ are called self-paced labels, where $\mathbf{y}_{i}$ is initialized by the original noisy pseudo label $\overline{\mathbf{y}}_{i}$ and acts as an online modified version for noise reduction. $\boldsymbol{\Phi}\left( \mathbf{x}_{i} \middle| \boldsymbol{W} \right)$ denotes a probability segmentation prediction of $\mathbf{x}_{i}$ by a discriminative function, i.e., the softmax layer of the CNN parameterized by $\boldsymbol{W}$. The cross-entropy loss between it and $\mathbf{y}_{i}$ is denoted by $L\left(\mathbf{y}_{i}, \boldsymbol{\Phi}\left(\mathbf{x}_{i} | \boldsymbol{W}\right)\right)$, and it is weighted by $\mathbf{v}_{i}$ as the first term in Eq.~\ref{SPL_Obj}. This involvement of the awareness of learning difficulty in model training improves the robustness against label noise.
Finally, we impose an $l_{2}$ regularization on $\boldsymbol{W}$ weighted by $\lambda$ to avoid model overfitting, as shown by the third term in Eq.~\ref{SPL_Obj}.

Objective function Eq.~\ref{SPL_Obj} can be minimized by alternating minimization strategy~\cite{spl}, where $\boldsymbol{W}$, $\boldsymbol{Y}$ and $\boldsymbol{V}$ are alternatively minimized one at a time, while the other two are fixed. The minimization in iteration $k$ consists of the following steps:
\begin{gather}
\label{updatingModel}
{\boldsymbol{W}^{(k)}=\underset{\boldsymbol{W}}{\operatorname{argmin}} \sum_{i}^{m} \mathbf{v}_{i}^{(k-1)} L\left(\mathbf{y}_{i}^{(k-1)}, \boldsymbol{\Phi}\left(\mathbf{x}_{i} | \boldsymbol{W}\right)\right)+\lambda\|\boldsymbol{W}\|_{2}}
\\
\label{updatingLabels}
{\boldsymbol{Y}^{(k)}=\underset{\boldsymbol{Y} \in\{0,1\}^{n \times m}}{\operatorname{argmin}} \sum_{i}^{m} L\left(\mathbf{y}_{i}, \boldsymbol{\Phi}\left(\mathbf{x}_{i} | \boldsymbol{W}^{(k)}\right)\right)}
\\
\label{updatingWeights}
\begin{split}
\boldsymbol{V}^{(k)} = \underset{\boldsymbol{V} \in[0,1]^{n \times m}} {\operatorname{argmin}} &\sum_{i}^{m} \mathbf{v}_{i} L\left(\mathbf{y}_{i}^{(k)}, \boldsymbol{\Phi}\left(\mathbf{x}_{i} | \boldsymbol{W}^{(k)}\right)\right)-\tau \left\|\mathbf{v}_{i}\right\|_{1}\\
&- \gamma \left\|\mathbf{v}_{i}\right\|_{2}
\end{split}
\end{gather}
The superscript $\left(k\right)$ represents the iteration index in alternating minimization. When $\boldsymbol{Y}$ and $\boldsymbol{V}$ are fixed, the optimization of $\boldsymbol{W}$ (Eq.~\ref{updatingModel}) is converted to the minimization of the sum of a weighted loss function and a regularization term, which can be typically solved by back-propagation.
When $\boldsymbol{W}$ and $\boldsymbol{V}$ are fixed, the optimization of $\boldsymbol{Y}$ (Eq.~\ref{updatingLabels}) is regarded as a model prediction problem and is solved by forward-propagation on the CNN with the optimal parameter $\boldsymbol{W}^{(k)}$ derived from Eq.~\ref{updatingModel}.
When $\boldsymbol{W}$ and $\boldsymbol{Y}$ are fixed, the optimization of $\boldsymbol{V}$ (Eq.~\ref{updatingWeights}) can be accomplished by SPLD algorithm \cite{diversity1}: pixels in each image are first sorted in ascending order of their losses and then assigned latent weights based on a threshold $\tau + \gamma/(\sqrt{o}+\sqrt{o-1})$ with respect to $\tau$, $\gamma$ and rank index $o$. Specifically, pixel $j$ in $\mathbf{x}_{i}$ with loss less than the threshold is selected as an easy pixel and is involved in training via assigning $v_{i,j}^{(k)} = 1$. Otherwise, it will not be included via assigning $v_{i,j}^{(k)} = 0$. The CNN is further trained in the next iteration $k+1$ using only the selected easy pixels by weighting the loss function with $\boldsymbol{V}^{(k)}$. Fig.~\ref{SPLFigure} shows some examples of self-paced labels and latent weights, i.e., in iteration $1$, $2$ and $3$ of alternating minimization.

\subsubsection{Sparse Annotation-based Manual Refinement} \label{sec_manual_refinement}

Naive self-paced learning scheme demonstrates the robustness for random label noise \cite{spl}, while it is hampered by systematic errors from vessel enhancement with specific biases to structures, such as thin and terminal branches with attenuated inflow of contrast agent.
Focusing only on easy pixels, the weighted loss function Eq.~\ref{updatingModel} may have a risk of ignoring systematic biases in noisy labels and thus lose crucial pixels for improving essential generalization capability. This leads to a poor segmentation model with suboptimal $\boldsymbol{W}$ and ruins the following alternating minimization steps. In particular, based on the current model, Eq.~\ref{updatingLabels} is dominated by easy pixels previously selected by the self-paced regularizer, maintaining and even amplifying systematic errors in the self-paced labels, as shown in Fig.~\ref{SPLFigure}. These systematic errors further misguide the model training in the next iteration, leading to irreversible performance deterioration at test time. Following the naive self-paced learning scheme, systematic errors are hardly explored and corrected if no auxiliary refinement strategy is applied.

\begin{figure}[t]
\centerline{\includegraphics[width=\columnwidth]{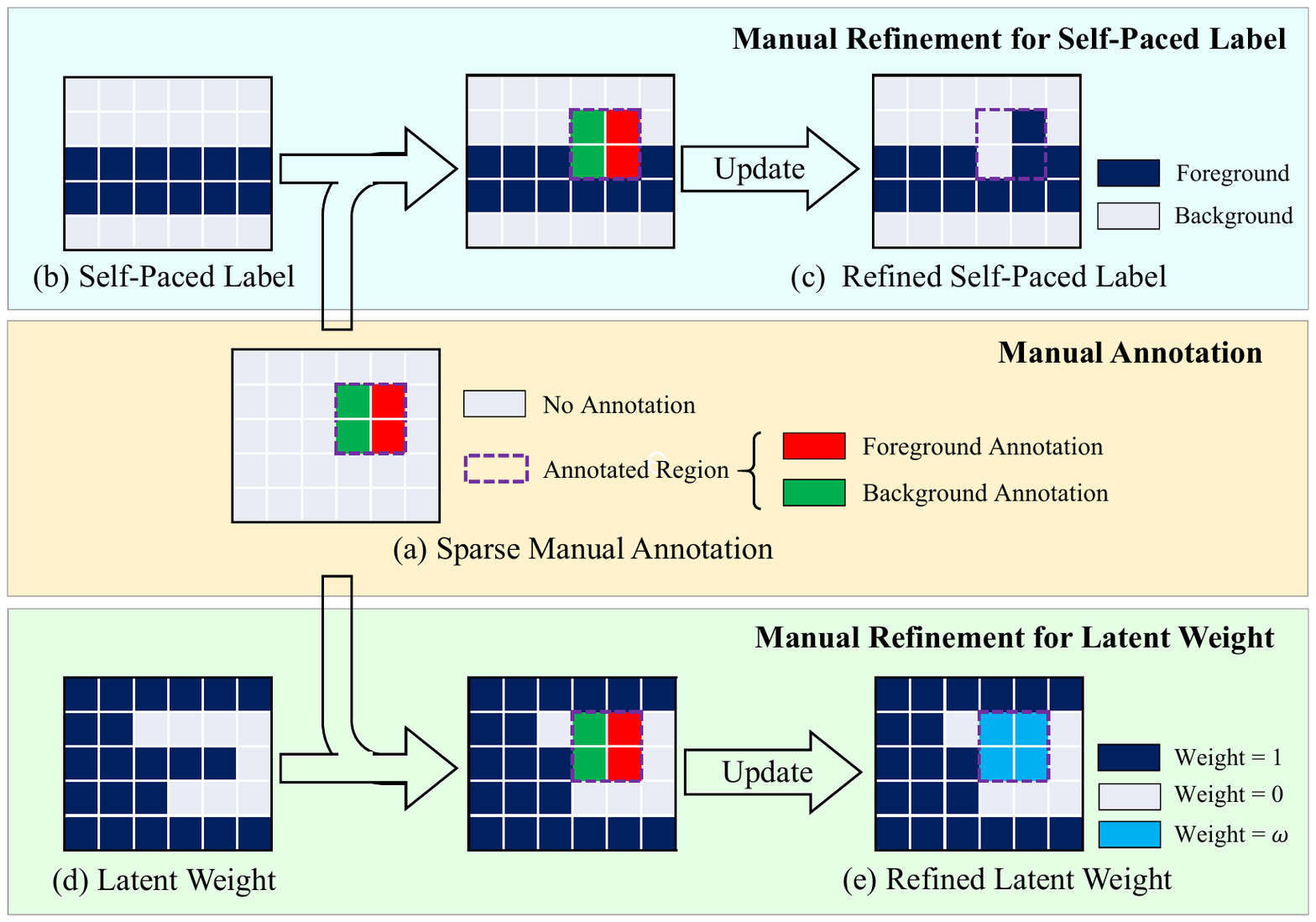}}
\caption{Illustration of sparse annotation-based manual refinement. (a) depicts the sparse annotations for a small number of pixels bounded by purple dash lines, where foreground and background annotations are colored in red and green, respectively. (b)$\rightarrow$(c) and (d)$\rightarrow$(e) depict the manual refinement for self-paced labels and latent weights. The refinement is performed only for the annotated region, where self-paced labels are updated with sparse annotations, and latent weights are updated with a constant $\omega$.}
\label{manualrefinement}
\end{figure}

Manual proofreading over the whole image is a practical way to detect and correct errors, while it leads to substantial labor costs for a large number of nonerror regions. Different from this labor-intensive process, we propose a cost-effective AR-SPL framework that performs an online local manual refinement based on sparse annotations to guide the naive self-paced learning, i.e., only a small number of valuable pixels with potentially incorrect labels are annotated and then manually refined in each iteration of alternating minimization during model training.
Specifically, given sparse annotations for a small portion of pixels in training image (Fig.~\ref{manualrefinement}(a)), manual refinement is performed only for the annotated region while the non-annotation region remains unchanged. In the annotated region, self-paced labels are updated with sparse annotations, as depicted in Fig.~\ref{manualrefinement}(b)$\rightarrow$(c). Moreover, the latent weights in the annotated region are increased to a constant $\omega>1$, as depicted in Fig.~\ref{manualrefinement}(d)$\rightarrow$(e), since sparse manual annotations are expected to more substantially impact model updating.
Owing to the paired manual refinements, the proposed AR-SPL provides a promising way to guide the naive self-paced learning to effectively overcome systematic label noise.
However, there is still a problem remaining: how to determine the sparse annotations that concentrate on a small number of valuable pixels with potentially incorrect labels, in order to minimize manual intervention as much as possible. This can be solved by suggestive annotation with model-vesselness uncertainty estimation, as introduced in Section~\ref{cap_suggestiveannotation}.

\subsubsection{Convergence Discussion}
Under some suitable assumptions, we show in Theorem~\ref{Convergence} that the proposed AR-SPL with manual refinement can converge stably to a stationary point. The theoretical proof is appended in Supplementary Materials. We show a detailed characterization of the stability of model parameters during training since it is the only relevant variable at test time. The experimental results in Section~\ref{cap_uncertaintyEstimation} also demonstrate that when manual annotations are gradually involved, the model achieves a higher segmentation performance and finally reaches a convergent result. In addition, following a similar argument, Theorem~\ref{Convergence} can be easily extended to imply the stability of latent weight and self-paced label, respectively.

\begin{theorem}\label{Convergence}
Denote the objective function in Eq.~\ref{SPL_Obj} as $\mathbb{F}\left(\boldsymbol{W},\boldsymbol{Y},\boldsymbol{V}\right)$. Let a unified optimization method, i.e., scholastic gradient descent with learning step $\alpha_{k}$, be used to solve Eq.~\ref{updatingModel},~\ref{updatingLabels},~\ref{updatingWeights} in iteration $k$ of alternating minimization. Let $\alpha_{k}$ satisfy $\sum_{k=0}^{\infty} \alpha_{k}=\infty$ and $\sum_{k=0}^{\infty} \alpha_{k}^{2}<\infty$. Then, $\lim _{k \rightarrow \infty} \mathbf{E}\left[\left\|\nabla_{\boldsymbol{W}} \mathbb{F}\left(\boldsymbol{W}^{(k)}, \boldsymbol{Y}^{(k)}, \boldsymbol{V}^{(k)}\right)\right\|_{2}\right]=0$.
\end{theorem}

\subsection{Suggestive Annotation with Model-Vesselness Uncertainty Estimation}\label{cap_suggestiveannotation}
In each iteration of AR-SPL, suggestive annotation is performed in two steps to achieve sparse annotations for manual refinement: i) intelligently querying a small number of valuable pixels for their labels from a large pool of unlabeled pixels; and ii) mixing the newly labeled pixels with the previously labeled ones.

\subsubsection{Batch-Mode Suggestive Annotation}
In principle, classical suggestive annotation chooses one single unlabeled sample each time~\cite{rep+uc,idnknow}, such as a pixel for the segmentation task, to query its label. However, it is not feasible enough in our work since one single queried pixel may not make a statistically significant impact on model updating. Moreover, labeling pixels in isolated positions one-by-one is intractable for the annotator as compared with labeling them in a localized image patch, i.e., a superpixel. Therefore, we perform suggestive annotation in batch mode at superpixel level for interaction efficiency.
For each training image, a small number of unlabeled superpixels with the highest uncertainties are queried. Then the annotator only needs to provide pixel-wise labels for these queried superpixels rather than the entire image. Therefore, these annotator-provided pixel-wise labels for queried superpixels are regarded as sparse annotations and used for manual refinement, as depicted in Fig.~\ref{manualrefinement}.

Formally, let image $\mathbf{x}_{i}$ be separated into a large number of superpixels that are denoted by a universal set $\boldsymbol{P}_{i}$. In iteration $k$  of AR-SPL, let $\boldsymbol{Q}_{i}^{(k)}$ denote a set of superpixels that need to be queried and annotated. Only $N_{b}$ superpixels (the query batch size) from the unlabeled pool with the highest uncertainties should be included in $\boldsymbol{Q}_{i}^{(k)}$ for sparse annotation:
\begin{equation}\label{querySelection}
    {\boldsymbol{Q}_{i}^{(k)}=\left\{q | q \in \boldsymbol{P}_{i}-\boldsymbol{Q}_{i}^{*}, ~U_{i, q}^{(k)} \geq \Gamma\left(N_{b}\right)\right\}}
\end{equation}
where $\boldsymbol{Q}_{i}^{*}=\boldsymbol{Q}_{i}^{(1)} \cup \boldsymbol{Q}_{i}^{(2)} \cup \cdots \cup \boldsymbol{Q}_{i}^{(k-1)}$ represents the labeled superpixels in the total previous $k-1$ iterations. $\boldsymbol{P}_{i}-\boldsymbol{Q}_{i}^{*}$ is the unlabeled pool obtained by excluding $\boldsymbol{Q}_{i}^{*}$ from $\boldsymbol{P}_{i}$, and we select superpixel $q$ from it to generate $\boldsymbol{Q}_{i}^{(k)}$.
Furthermore, $U_{i, q}^{(k)}$ measures the uncertainty for $q$ in $\mathbf{x}_{i}$. $\Gamma\left(N_{b}\right)$ is the $N_{b}$-$th$ highest uncertainty among all superpixels. The selected $q$ should also have the top $N_{b}$ uncertainty in $\mathbf{x}_{i}$, which satisfies $U_{i, q}^{(k)} \geq \Gamma\left(N_{b}\right)$.
Finally, all pixels in $\boldsymbol{Q}_{i}^{(k)}$ are labeled as sparse annotations and then added to the previously labeled $\boldsymbol{Q}_{i}^{*}$ for manual refinement in AR-SPL, in order to avoid catastrophic forgetting \cite{Li2018LearningWF} of CNNs.
Uncertainty estimation $U_{i, q}^{(k)}$ is regarded as a key for the selection of $\boldsymbol{Q}_{i}^{(k)}$ and will be described in the next subsection.

\subsubsection{Model-Vesselness Uncertainty Estimation}
Towards manual refinement specific for errors in current self-paced labels, uncertainty estimation is desired to indicate potential mis-segmentations \cite{testtimeAug}, which can be achieved with a widely used model uncertainty using Monte Carlo sampling with dropout (MCDO) \cite{modeluncertainty}. For this, in iteration $k$ of AR-SPL, we first activate the dropout operation in model inference and perform $D$ times forward-propagations, leading to $D$-fold binary prediction results. Then, the model posterior expectation is obtained by averaging over them:
\begin{equation}\label{MCDO}
    {E_{i, j}^{(k)}=\frac{1}{D} \sum_{d}^{D} \tilde{y}_{i, j}^{(k)}}
\end{equation}
where $E_{i, j}^{(k)}$ denotes the model expectation for $x_{i,j}$ (pixel $j$ in $\mathbf{x}_{i}$). ${\tilde{y}_{i, j}^{(k)}} = \boldsymbol{\tilde{\boldsymbol{\Phi}}}\left( x_{i,j} \middle| \boldsymbol{W}_{d}^{(k)} \right)$ is the binary prediction for $x_{i,j}$, where $\boldsymbol{W}_{d}^{(k)}$ denotes the model parameter $\boldsymbol{W}^{(k)}$ after applying dropout in MCDO pass $d$. Furthermore, model uncertainty $M_{i,j}^{(k)}$ is estimated as the entropy over $E_{i, j}^{(k)}$:
\begin{equation}\label{MC_equation}
    {M_{i,j}^{(k)} = - \frac{1}{{\check{Z}}_{i}}E_{i,j}^{(k)}{\log E_{i,j}^{(k)}}}
\end{equation}
where ${\check{Z}}_{i} = {\max\limits_{j}{E_{i,j}^{(k)}{\log E_{i,j}^{(k)}}}}$ is the normalization parameter to make $M_{i,j}^{(k)}$ range from $0$ to $1$.

Despite the strong relationship with potential mis-segmentations, model uncertainty leads to redundant queries which limit the intervention efficiency, especially for the model training at early stages, due to the absence of geometric features of coronary arteries. Vesselness measure generated from vessel enhancement is regarded as a customized geometric feature for vascular structures, which can be used to define vesselness uncertainty $G_{i,j}$:
\begin{equation}\label{VC_equation}
    {G_{i,j} = - \frac{1}{{\hat{Z}}_{i}}s_{i,j}\left( {1 - s_{i,j}} \right)}
\end{equation}
where $s_{i,j}$ represents the vesselness measure from Section \ref{PseudoLabelGeneration}, and ${\hat{Z}}_{i} = {\max\limits_{j}{s_{i,j}\left( {1 - s_{i,j}} \right)}}$ is the normalization parameter. Eq.~\ref{VC_equation} is formulated as a quadratic function \cite{costEffective} rather than a widely used entropy term, considering the distribution difference between $s_{i,j}$ and $E_{i,j}^{(k)}$.

To leverage the complementary strengths of these two uncertainty estimations, we propose a novel model-vesselness uncertainty that is a combination of them. It is formulated at superpixel level for the batch-mode suggestive annotation in iteration $k$ of AR-SPL:
\begin{equation}\label{MVU_equation}
    {U_{i,q}^{(k)} = \frac{1}{N_{q}}{\sum\limits_{j \in q}{\operatorname{max}\left({\eta G_{i,j},\left( {1 - \eta} \right)M_{i,j}^{(k)}} \right)}}}
\end{equation}
where $U_{i,q}^{(k)}$ denotes the proposed model-vesselness uncertainty for superpixel $q$ in $\mathbf{x}_{i}$, and $N_{q}$ represents the total number of pixels in it. This hybrid uncertainty calculates a weighted maximization of model uncertainty and vesselness uncertainty and then averages over all pixels in $q$.
The weight $\eta$ controls a tradeoff between them, which requires an elaborate design for the best combination. Specifically, vesselness uncertainty provides a context-aware cue for suggestive annotation. It exhibits more advantages for an early training stage where model uncertainty is unreliable due to the inaccurate predictions based on a coarse segmentation model. However, vesselness uncertainty cannot discover which pixels are actually essential for further model fine-tuning, leading to decreased convergent performance. In contrast, model uncertainty indicates ambiguous regions with respect to the segmentation model. It allows for better exploration of potentially incorrect labels \cite{testtimeAug}, leading to an accurate and stable segmentation performance for model convergence. Motivated by these observations, we design a dynamic time-dependent weight $\eta$, regarded as a soft switching strategy between these two uncertainties during the entire training process:
\begin{equation}\label{tradeoff}
    {\eta} =
    \begin{cases}
        {1 - \theta\left( {k - 1} \right)}, & {k < 1 + \left( {1/\theta} \right)}\\
        0, & \operatorname{otherwise}
    \end{cases}
\end{equation}
where $\theta$ denotes the decay rate and $k$ is the training iteration index. With this dynamic tradeoff, the proposed model-vesselness uncertainty emphasizes vesselness uncertainty more in the early training stage for a fast performance improvement, while biasing to model uncertainty in the later stage for an accurate convergent performance.

\section{Experiments and Results}

\subsection{Dataset and Evaluation Metrics}

We collected 191 clinical XA sequences of 30 patients with frame rate 15 $fps$, frame size ${512 \times 512}$ and pixel size ${0.3 \times 0.3}$ mm$^2$ using a Philips UNIQ FD10 C-arm system from Peking Union Medical College Hospital in China. Among them, multiple sequences were acquired for each patient from multiple viewing angles to overcome the foreshortening of angiographic projection.
XA sequences recorded the whole vessel angiography procedure from the inflow to the wash-out of the injected contrast agent, with total frames 41 to 74. During the angiography procedure, only the key frame depicted the entire structure of coronary tree by filling the vessel lumen with contrast agent~\cite{syeda2010automatic}. Therefore, we selected one contrast-filled key frame from each sequence and obtained a total of 191 XAs of 30 patients for the following segmentation experiments. Data splitting was performed at the patient level. We used 112 XAs of 17 patients for training, 25 XAs of 4 patients for validation and 54 XAs of 9 patients for testing.

We developed a PyQT GUI for elaborate vessel annotation. All of the XAs can be enlarged up to 5$\times$ for clear visualization of even thin branches, and vessel regions were annotated by a laser mouse.
The annotator was an experienced researcher who can accurately identify coronary arteries, and the annotation quality was further checked by an expert radiologist for PCI surgery.
The annotator not only labeled superpixels that were queried in training images, but also provided the vessel ground truth for validation and testing images.
To quantitatively evaluate the segmentation performance, we measured recall, precision and dice score (which is equal to F1-score):
\begin{gather}
    \text{Recall} = TP/\left( {TP + FN} \right)
    \\
    \text{Precision} = TP/\left( {TP + FP} \right)
    \\
    \text{Dice Score} = 2TP/\left( {2TP + FN + FP} \right)
\end{gather}
where $TP$, $FN$ and $FP$ are the numbers of true positives, false negatives and false positives of segmentation results, respectively. Considering the high class imbalance in XA, dice score provides a relatively more comprehensive evaluation than recall and precision.

\begin{figure*}[t]
\centerline{\includegraphics[width=7.3in]{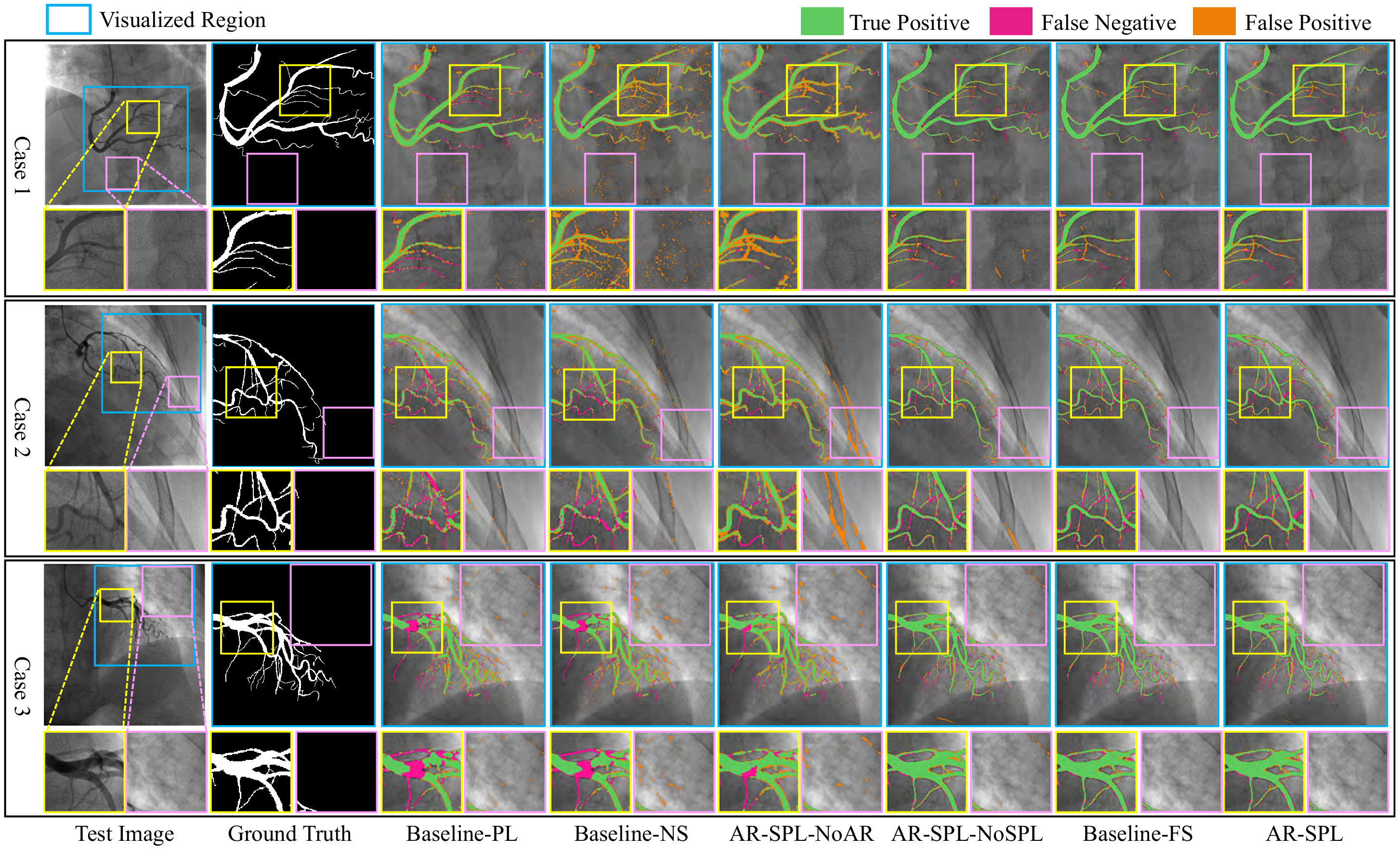}}
\caption{Three examples of segmentation results in blue bounding boxes obtained by different counterparts in the ablation study. The true positives, false negatives and false positives are visualized in green, red and orange, respectively. We zoom in on the challenging regions bounded by yellow and pink boxes, and append them below each case to highlight the segmentation details.}
\label{AblationStudy}
\end{figure*}

\subsection{Implementation Details} \label{Implementation}
The proposed weakly supervised learning framework was implemented in TensorFlow\footnote{https://www.tensorflow.org} with a 4-core 2.6GHz Intel Xeon Silver processor, an NVIDIA Titan X (Pascal) GPU and 128 GB RAM.
\vskip0.5ex
\noindent
\textbf{Pseudo Label Generation.}
We followed the suggestions from \cite{vcrpca} to generate pseudo labels, e.g., a structural disk element with 20 pixel diameter and regularization parameter $\xi = 0.8/\sqrt{n}$, where $n$ is the number of pixels in a training image.
\vskip0.5ex
\noindent
\textbf{Annotation-Refining Self-Paced Learning.} Without loss of generality, we chose the widely used U-Net \cite{unet} as the segmentation model. Dropout with rate $0.2$ was used in the expanding path of U-Net to reduce overfitting and acted as Monte Carlo sampling for model uncertainty estimation. Note that the dropout was applied only for model training Eq.~\ref{updatingModel} and MCDO Eq.~\ref{MCDO} rather than in the testing process, where it would cause multiplied inference time.
Sparse annotation-based manual refinement was performed in each iteration of AR-SPL and finally stopped at iteration 13 when a convergence threshold was satisfied, i.e., the increment of dice score on the validation dataset is less than 0.05\%.
During the training process, we first initialized the self-paced label with the noisy pseudo label obtained from vessel enhancement. Moreover, a soft threshold was empirically imposed on vesselness map to initialize the latent weight $\mathbf{v}_{i} = {\max\left( {4\left| {\mathbf{s}_{i} - 0.5} \right|,1} \right)}$. Then, Eq.~\ref{updatingModel} was optimized with $\lambda = 10^{- 4}$ by stochastic gradient descent with momentum 0.9, batch size 16, maximal number of iterations 5000 and initial learning rate 0.01 that was decayed exponentially with power 0.9.
For the optimization of Eq.~\ref{updatingWeights}, we synchronously logarithmically decreased $\tau$ and $\gamma$ from different initial values $\tau_{0}$ and $\gamma_{0}$ with the same rate $\mu$, i.e. $\tau = - {\log\left( {\tau_{0} - \mu k} \right)}$ and $\gamma = - {\log\left( {\gamma_{0} - \mu k} \right)}$, so as to maintain the stability and scalability of the selection threshold for cross-entropy loss. The analysis of parameter sensitivity is appended in Supplementary Materials. It suggests $\tau_{0}=0.75$, $\gamma_{0}=0.20$ and $\mu=0.01$ based on a grid search for the best performance on the validation dataset. Following a similar grid search manner, we chose $\omega = 5$ to emphasize manual refinement for latent weight.
\vskip0.5ex
\noindent
\textbf{Suggestive Annotation with Model-Vesselness Uncertainty Estimation.} We utilized the SLIC algorithm \cite{SLIC} to separate each training image into 3000 superpixels.
The query batch size $N_{b}$ was set to 8 due to its best tradeoff between human-related annotation cost and model performance, as shown in Supplementary Materials.
Model uncertainty was estimated by 20-fold MCDO and then dynamically combined with vesselness uncertainty with decay rate $\theta=0.4$, which was selected based on a grid search for the best validation performance.

\subsection{Ablation Study}\label{Cap_AblationStudy}

In this section, we conduct an ablation study to investigate the contribution of each component in the proposed AR-SPL: the pseudo label generation, the self-paced learning scheme and the spare annotation-based manual refinement. We compare three baselines and several model variants that use different components in AR-SPL.
\begin{itemize}
    \setlength{\itemsep}{0pt}
    \setlength{\parsep}{0pt}
    \setlength{\parskip}{2pt}
    \item Pseudo label prediction, i.e., a baseline that predicts pseudo labels (\textbf{Baseline-PL}) based on vessel enhancement for test images. These pseudo labels are directly regarded as the final segmentation results without further fed to deep learning-based training framework.
    \item Noisy supervision-based training framework \cite{BaselineNS}, i.e., a baseline that treats pseudo labels as noisy supervision (\textbf{Baseline-NS}) and then learns directly from them without further refinement strategies.
    \item Automatic modification of pseudo labels based on the naive self-paced learning scheme without manual refinement, i.e., deactivation of sparse annotation-based manual refinement in AR-SPL (\textbf{AR-SPL-NoAR}).
    \item Manual substitution of pseudo labels with sparse annotations that are suggested by model-vesselness uncertainty. Only these sparse annotations are used in model training, which is accomplished without the self-paced learning for exploiting pseudo labels on the fly, i.e., deactivation of the self-paced learning scheme in AR-SPL (\textbf{AR-SPL-NoSPL})\footnote{For a fair comparison, to avoid additional annotations for initialization~\cite{ASPL,repu}, we only use noisy pseudo labels as a warm-start to obtain the initial segmentation model and uncertainty map.}.
    \item Fully supervision-based training framework \cite{cnnvesselseg1}, i.e., a baseline that learns from fully supervision (\textbf{Baseline-FS}) with manual annotations for all pixels in each training image.
\end{itemize}

\begin{table*}[t]
    \centering
    \caption{Quantitative evaluations for training cost and segmentation performances of all methods that use different components in the ablation study, such as the fully supervision (FS), the pseudo label generation (PLG), the naive self-paced learning scheme (SPL) and the sparse annotation-based manual refinement (AR). The best performance is highlighted in bold, and its comparable performance is denoted by superscript $^\textbf{c}$ based on a two-sided Wilcoxon signed-rank test ($p$-value$>$0.05).}
    \setlength{\tabcolsep}{-2.3 mm}{
    \begin{tabular}{lcccccccccc}
    \specialrule{0.05em}{0pt}{1pt}
    \multirow{2}{*}{} &
    \multicolumn{4}{c}{\raisebox{2pt}{\phantom{M}} \footnotesize Component \raisebox{-2pt}{\phantom{M}}} &
    \multicolumn{3}{c}{\raisebox{2pt}{\phantom{M}} \footnotesize Training Cost \raisebox{-2pt}{\phantom{M}}} &
    \multicolumn{3}{c}{\raisebox{2pt}{\phantom{M}} \footnotesize Segmentation Performance \raisebox{-2pt}{\phantom{M}}}\\\specialrule{0em}{0pt}{0pt}
    \cmidrule(r){2-5} \cmidrule(r){6-8} \cmidrule(r){9-11}
    \noalign{\smallskip}
    \raisebox{0pt}{\phantom{M}} \raisebox{-1pt}{\phantom{M}} &
    \raisebox{0pt}{\phantom{M}} \begin{tabular}[c]{@{}c@{}}\footnotesize FS \end{tabular} \raisebox{-1pt}{\phantom{M}} &
    \raisebox{0pt}{\phantom{M}} \begin{tabular}[c]{@{}c@{}}\footnotesize PLG \end{tabular} \raisebox{-1pt}{\phantom{M}} &
    \raisebox{0pt}{\phantom{M}} \begin{tabular}[c]{@{}c@{}}\footnotesize SPL \end{tabular} \raisebox{-1pt}{\phantom{M}} &
    \raisebox{0pt}{\phantom{M}} \begin{tabular}[c]{@{}c@{}}\footnotesize AR \end{tabular} \raisebox{-1pt}{\phantom{M}} &
    \raisebox{0pt}{\phantom{M}} \begin{tabular}[c]{@{}c@{}}\footnotesize \# Annotated\\\footnotesize Superpixels ($k$)\end{tabular} \raisebox{-1pt}{\phantom{M}}  &
    \raisebox{0pt}{\phantom{M}} \begin{tabular}[c]{@{}c@{}}\footnotesize Annotation\\\footnotesize Time ($h$)\end{tabular} \raisebox{-1pt}{\phantom{M}} &
    \raisebox{0pt}{\phantom{M}} \begin{tabular}[c]{@{}c@{}}\footnotesize Optimization\\\footnotesize Time ($h$)\end{tabular} \raisebox{-1pt}{\phantom{M}} &
    \raisebox{0pt}{\phantom{M}} \begin{tabular}[c]{@{}c@{}}\footnotesize Recall\\(\%)\end{tabular} \raisebox{-1pt}{\phantom{M}} &
    \raisebox{0pt}{\phantom{M}} \begin{tabular}[c]{@{}c@{}}\footnotesize Precision\\(\%)\end{tabular} \raisebox{-1pt}{\phantom{M}} &
    \raisebox{0pt}{\phantom{M}} \begin{tabular}[c]{@{}c@{}}\footnotesize Dice\\(\%)\end{tabular} \raisebox{-1pt}{\phantom{M}}\\ \specialrule{0.05em}{2pt}{1pt}
    
   \raisebox{2pt}{\phantom{M}} \footnotesize Baseline-PL \raisebox{-2pt}{\phantom{M}} &
    \raisebox{2pt}{\phantom{M}} \raisebox{-2pt}{\phantom{M}} &
    \raisebox{2pt}{\phantom{M}} \CheckmarkBold \raisebox{-2pt}{\phantom{M}} &
    \raisebox{2pt}{\phantom{M}} \raisebox{-2pt}{\phantom{M}} &
    \raisebox{2pt}{\phantom{M}} \raisebox{-2pt}{\phantom{M}} &
    \raisebox{2pt}{\phantom{M}} \footnotesize 0 \raisebox{-2pt}{\phantom{M}} &
    \raisebox{2pt}{\phantom{M}} \footnotesize 0 \raisebox{-2pt}{\phantom{M}} &
    \raisebox{2pt}{\phantom{M}} \footnotesize 0 \raisebox{-2pt}{\phantom{M}} &
    \raisebox{2pt}{\phantom{M}} \footnotesize 77.37$\pm$9.58 \raisebox{-2pt}{\phantom{M}} &
    \raisebox{2pt}{\phantom{M}} \footnotesize 56.57$\pm$11.48 \raisebox{-2pt}{\phantom{M}} &
    \raisebox{2pt}{\phantom{M}} \footnotesize 61.64$\pm$10.42 \raisebox{-2pt}{\phantom{M}}\\\specialrule{0em}{0.5pt}{0.5pt}

    \raisebox{2pt}{\phantom{M}} \footnotesize Baseline-NS \raisebox{-2pt}{\phantom{M}} &
    \raisebox{2pt}{\phantom{M}} \raisebox{-2pt}{\phantom{M}} &
    \raisebox{2pt}{\phantom{M}} \CheckmarkBold \raisebox{-2pt}{\phantom{M}} &
    \raisebox{2pt}{\phantom{M}} \raisebox{-2pt}{\phantom{M}} &
    \raisebox{2pt}{\phantom{M}} \raisebox{-2pt}{\phantom{M}} &
    \raisebox{2pt}{\phantom{M}} \footnotesize 0 \raisebox{-2pt}{\phantom{M}} &
    \raisebox{2pt}{\phantom{M}} \footnotesize 0 \raisebox{-2pt}{\phantom{M}} &
    \raisebox{2pt}{\phantom{M}} \footnotesize 3.26 \raisebox{-2pt}{\phantom{M}} &
    \raisebox{2pt}{\phantom{M}} \footnotesize 83.56$\pm$7.28 \raisebox{-2pt}{\phantom{M}} &
    \raisebox{2pt}{\phantom{M}} \footnotesize 43.60$\pm$10.86 \raisebox{-2pt}{\phantom{M}} &
    \raisebox{2pt}{\phantom{M}} \footnotesize 56.19$\pm$9.60 \raisebox{-2pt}{\phantom{M}}\\\specialrule{0em}{0.5pt}{0.5pt}

    \raisebox{2pt}{\phantom{M}} \footnotesize AR-SPL-NoAR \raisebox{-2pt}{\phantom{M}} &
    \raisebox{2pt}{\phantom{M}} \raisebox{-2pt}{\phantom{M}} &
    \raisebox{2pt}{\phantom{M}} \CheckmarkBold \raisebox{-2pt}{\phantom{M}} &
    \raisebox{2pt}{\phantom{M}} \CheckmarkBold \raisebox{-2pt}{\phantom{M}} &
    \raisebox{2pt}{\phantom{M}} \raisebox{-2pt}{\phantom{M}} &
    \raisebox{2pt}{\phantom{M}} \footnotesize 0 \raisebox{-2pt}{\phantom{M}} &
    \raisebox{2pt}{\phantom{M}} \footnotesize 0 \raisebox{-2pt}{\phantom{M}} &
    \raisebox{2pt}{\phantom{M}} \footnotesize 11.57 \raisebox{-2pt}{\phantom{M}} &
    \raisebox{2pt}{\phantom{M}} \footnotesize \textbf{86.86$\pm$5.84} \raisebox{-2pt}{\phantom{M}} &
    \raisebox{2pt}{\phantom{M}} \footnotesize 49.89$\pm$7.27 \raisebox{-2pt}{\phantom{M}} &
    \raisebox{2pt}{\phantom{M}} \footnotesize 63.09$\pm$6.58 \raisebox{-2pt}{\phantom{M}}\\\specialrule{0em}{0.5pt}{0.5pt}

    \raisebox{2pt}{\phantom{M}} \footnotesize AR-SPL-NoSPL \raisebox{-2pt}{\phantom{M}} &
    \raisebox{2pt}{\phantom{M}} \raisebox{-2pt}{\phantom{M}} &
    \raisebox{2pt}{\phantom{M}} \CheckmarkBold \raisebox{-2pt}{\phantom{M}} &
    \raisebox{2pt}{\phantom{M}} \raisebox{-2pt}{\phantom{M}} &
    \raisebox{2pt}{\phantom{M}} \CheckmarkBold \raisebox{-2pt}{\phantom{M}} &
    \raisebox{2pt}{\phantom{M}} \footnotesize 11.64 \raisebox{-2pt}{\phantom{M}} &
    \raisebox{2pt}{\phantom{M}} \footnotesize 16.32 \raisebox{-2pt}{\phantom{M}} &
    \raisebox{2pt}{\phantom{M}} \footnotesize 11.74 \raisebox{-2pt}{\phantom{M}} &
    \raisebox{2pt}{\phantom{M}} \footnotesize 82.07$\pm$3.93 \raisebox{-2pt}{\phantom{M}} &
    \raisebox{2pt}{\phantom{M}} \footnotesize 81.53$\pm$5.49 \raisebox{-2pt}{\phantom{M}} &
    \raisebox{2pt}{\phantom{M}} \footnotesize 81.72$\pm$4.14 \raisebox{-2pt}{\phantom{M}}\\\specialrule{0em}{0.5pt}{0.5pt}

    \raisebox{2pt}{\phantom{M}} \footnotesize Baseline-FS \raisebox{-2pt}{\phantom{M}} &
    \raisebox{2pt}{\phantom{M}} \CheckmarkBold \raisebox{-2pt}{\phantom{M}} &
    \raisebox{2pt}{\phantom{M}} \raisebox{-2pt}{\phantom{M}} &
    \raisebox{2pt}{\phantom{M}} \raisebox{-2pt}{\phantom{M}} &
    \raisebox{2pt}{\phantom{M}} \raisebox{-2pt}{\phantom{M}} &
    \raisebox{2pt}{\phantom{M}} \footnotesize 336.00 \raisebox{-2pt}{\phantom{M}} &
    \raisebox{2pt}{\phantom{M}} \footnotesize 65.11 \raisebox{-2pt}{\phantom{M}} &
    \raisebox{2pt}{\phantom{M}} \footnotesize 3.24 \raisebox{-2pt}{\phantom{M}} &
    \raisebox{2pt}{\phantom{M}} \footnotesize 81.44$\pm$4.26 \raisebox{-2pt}{\phantom{M}} &
    \raisebox{2pt}{\phantom{M}} \footnotesize 82.58$\pm$5.35$^{\textbf{c}}$ \raisebox{-2pt}{\phantom{M}} &
    \raisebox{2pt}{\phantom{M}} \footnotesize 82.01$\pm$4.21$^{\textbf{c}}$ \raisebox{-2pt}{\phantom{M}}\\\specialrule{0em}{0.5pt}{0.5pt}

    \raisebox{2pt}{\phantom{M}} \footnotesize AR-SPL \raisebox{-2pt}{\phantom{M}} &
    \raisebox{2pt}{\phantom{M}} \raisebox{-2pt}{\phantom{M}} &
    \raisebox{2pt}{\phantom{M}} \CheckmarkBold \raisebox{-2pt}{\phantom{M}} &
    \raisebox{2pt}{\phantom{M}} \CheckmarkBold \raisebox{-2pt}{\phantom{M}} &
    \raisebox{2pt}{\phantom{M}} \CheckmarkBold \raisebox{-2pt}{\phantom{M}} &
    \raisebox{2pt}{\phantom{M}} \footnotesize 11.64 \raisebox{-2pt}{\phantom{M}} &
    \raisebox{2pt}{\phantom{M}} \footnotesize 16.16 \raisebox{-2pt}{\phantom{M}} &
    \raisebox{2pt}{\phantom{M}} \footnotesize 11.81 \raisebox{-2pt}{\phantom{M}} &
    \raisebox{2pt}{\phantom{M}} \footnotesize 81.64$\pm$4.11 \raisebox{-2pt}{\phantom{M}} &
    \raisebox{2pt}{\phantom{M}} \footnotesize \textbf{82.69$\pm$5.31} \raisebox{-2pt}{\phantom{M}} &
    \raisebox{2pt}{\phantom{M}} \footnotesize \textbf{82.09$\pm$4.08} \raisebox{-2pt}{\phantom{M}}\\\specialrule{0.05em}{0.5pt}{0.5pt}
    \end{tabular}}
    \label{abalationStudyEvaluation}
\end{table*}

Fig.~\ref{AblationStudy} visualizes the segmentation results on test images obtained by different methods in the ablation study, zooming in on the challenging regions for vessels (yellow bounding boxes) and background with potential disturbance (pink bounding boxes).
Baseline-PL predicts pseudo label as the segmentation result, which exhibits the obvious false negatives for thin vessel branches and bifurcation points (in the yellow bounding boxes), as well as the false positives scattered in semi-transparent background structures (in the pink bounding boxes). This poor performance indicates that pseudo labels contain noise with systematic biases, challenging the other components in AR-SPL for learning an accurate segmentation model. For example, Baseline-NS directly learns from pseudo labels while shows even worse performance than Baseline-PL due to the overfitting on label noise. AR-SPL-NoAR and AR-SPL-NoSPL exhibit limited noise robustness, where false negatives are reduced as compared to Baseline-PL and Baseline-NS, yet background disturbance seems to be intractable with false positives even amplified especially for curvilinear background structures, such as ribs, sternums and vertebrae. Owing to the incorporation of all proposed components, AR-SPL can effectively avoid the model training being corrupted by noise in pseudo labels, leading to the accurate extraction of contrast-filled coronary arteries as highlighted in yellow bounding boxes and the least false positives for background disturbance in pink bounding boxes. Its performance is significantly better than other counterparts that use one single component, and even is visually similar to Baseline-FS that relies on tedious fully manual annotation.

We show in Table~\ref{abalationStudyEvaluation} the quantitative evaluations of training cost and segmentation performance. Training cost consists of the human-related annotation cost (i.e., the number of annotated superpixels and their required annotation time for all images in the entire training process) and the human-free optimization cost (i.e., the optimization time for updating operations in alternating minimization Eq.~\ref{updatingModel},~\ref{updatingLabels},~\ref{updatingWeights}).
Baseline-PL is an unsupervised method based on pseudo label generation without training process, which thus requires no annotation cost and optimization cost for model training. Its low dice score demonstrates that pseudo labels are highly noisy and would challenge other components in AR-SPL. Specifically, Baseline-NS shows even worse performance than Baseline-PL, indicating the overfitting nature of CNNs for label noise. AR-SPL-NoAR is completely free of manual annotation while has limited improvement over Baseline-PL and Baseline-NS, which highlights that using only the naive self-paced learning scheme tends to be limited by highly noisy pseudo labels. It increases the recall yet has a much lower precision due to the lack of discrimination capability for false-positive ambiguous structures. AR-SPL-NoSPL prominently promotes the segmentation performance over Baseline-PL and Baseline-NS via abandoning all pseudo labels and learning only from sparse annotations suggested by uncertainty estimation. It requires the similar training cost to AR-SPL yet has an inferior segmentation performance due to the absence of the self-paced learning for leveraging potentially clean pseudo labels with low learning difficulties on the fly. 
Compared with other counterparts, AR-SPL integrates all proposed components and achieves the best segmentation performance, which is statistically comparable to Baseline-FS and even exhibits a slight superiority without significant difference. This is because that when the uncertainty is used to guide sparse annotation-based manual refinement, more attention with a large weight $\omega$ would be paid to these difficult regions during model training. Moreover, compared with Baseline-FS, AR-SPL significantly reduces the annotation cost, i.e., only 24.82\% annotation time is required to label 3.46\% image regions. Although it costs more optimization time for alternating minimization, such increase of optimization time is fairly slight considering the largely reduced annotation time, and it is also acceptable since no human-related labor is involved in optimization procedure. These advantages of AR-SPL demonstrate the contribution of each component for the safety (without performance deterioration) and efficiency (with minimal annotation cost) when dealing with noisy pseudo labels.

\begin{figure}[t]
\centerline{\includegraphics[width=3in]{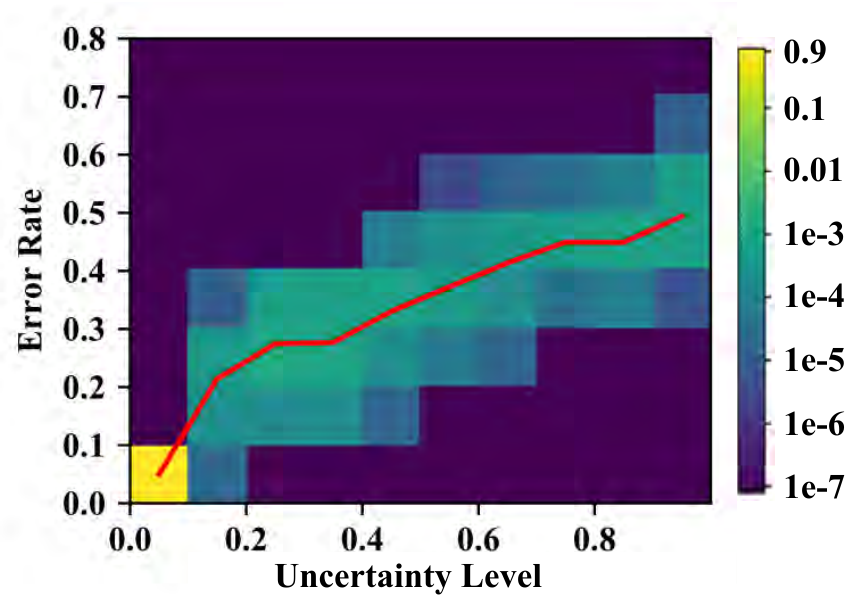}}
\caption{Normalized joint histogram of uncertainty and error rate. The average error rate is depicted as a function of uncertainty by the red curve.}
\label{error_rate}
\end{figure}

Manual refinement is performed by precisely annotating only superpixels with high uncertainties, rather than extensively proofreading the entire image which is a slow and labor-intensive process.
To investigate the feasibility of this uncertainty-based suggestive annotation for local manual refinement, we show in Fig.~\ref{error_rate} the relationship between the adopted model-vesselness uncertainty and segmentation error. At different uncertainty levels, we measure error rates of segmentation results for all images, obtaining a normalized joint histogram of uncertainty and segmentation error rate. Then, we also calculate the average error rate with respect to different uncertainty levels and present the error rate as a function of uncertainty, i.e., the red curve in Fig.~\ref{error_rate}. The results demonstrate that the majority of pixels have correct segmentation predictions (low error rate) with low uncertainties. The error rate becomes gradually higher with the increase of uncertainty, indicating that the segmentation error can be captured by a higher uncertainty value. Therefore, the uncertainty-based suggestive annotation enables a reliable local manual refinement for potential segmentation errors, providing a cost-effective alternative for whole-image proofreading.

\subsection{Comparison with Other Uncertainty Estimations in Suggestive Annotation}\label{cap_uncertaintyEstimation}

To demonstrate the advantage of the proposed adaptive model-vesselness uncertainty (\textbf{MVU-ada}) for a more efficient manual intervention, we compare it with other uncertainty estimations in the same AR-SPL framework:
\begin{itemize}
    \setlength{\itemsep}{0pt}
    \setlength{\parsep}{0pt}
    \setlength{\parskip}{2pt}
    \item Model uncertainty (\textbf{MU}) \cite{modeluncertainty}, which is estimated by MCDO, as shown in Eq.~\ref{MC_equation}.
    \item Vesselness uncertainty (\textbf{VU}), which is derived based on the vesselness map generated from vessel enhancement, as shown in Eq.~\ref{VC_equation}.
    \item Model-vesselness uncertainty with a fix tradeoff (\textbf{MVU-fix}), which is a hybrid uncertainty estimated as the weighted maximization of model uncertainty and vesselness uncertainty, as shown in Eq.~\ref{MVU_equation}. Unlike MVU-ada, it adopts a fixed weight $\eta=0.2$ rather than the dynamic weight formulated in Eq.~\ref{tradeoff}, which is chosen by a grid search for the best validation performance.
\end{itemize}

\begin{figure}[t]
\centerline{\includegraphics[width=\columnwidth]{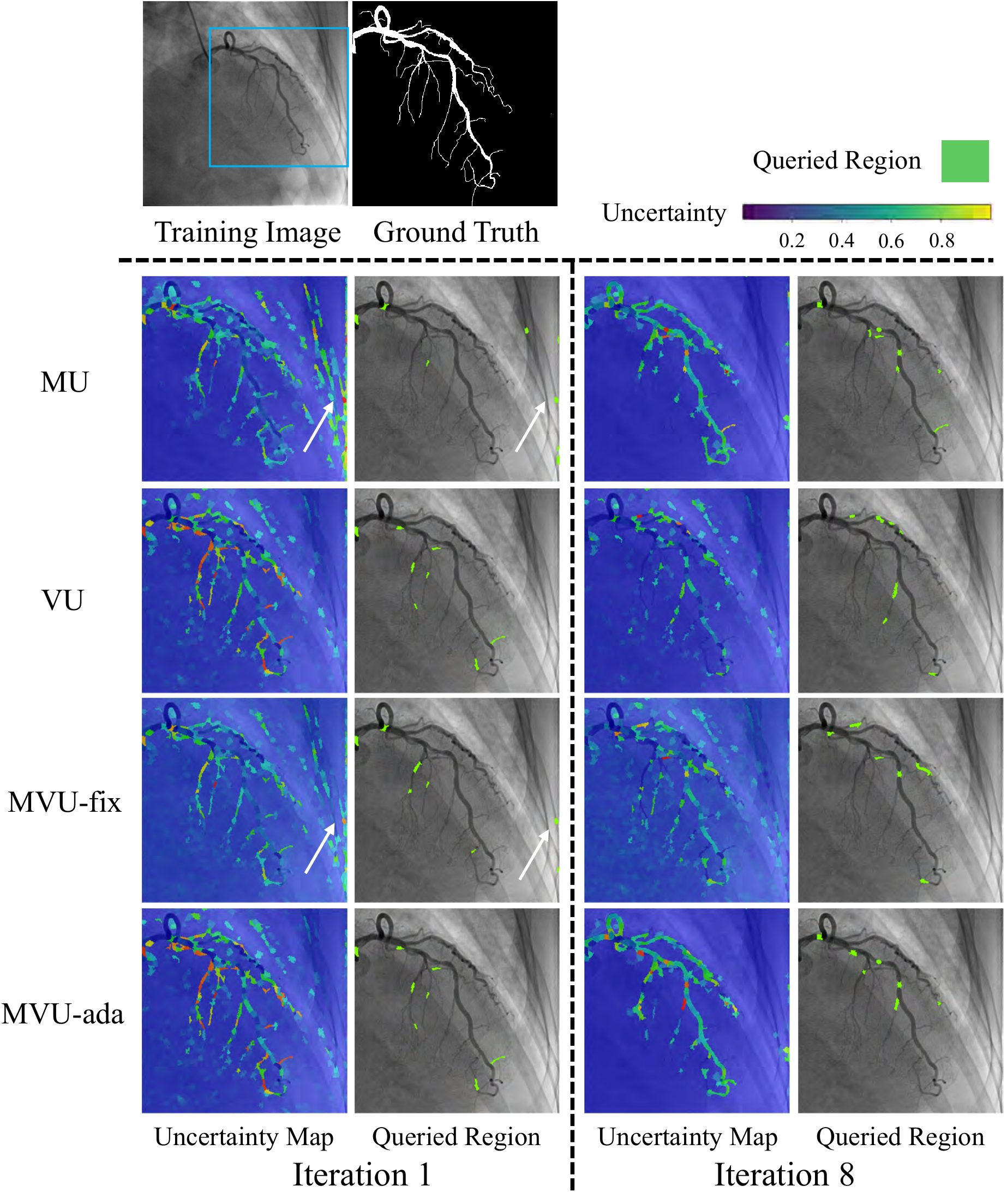}}
\caption{Visual comparison of different uncertainty maps and their corresponding queried regions in iterations 1 and 8 of AR-SPL. The uncertainty maps are visualized by heatmaps, and the queried regions are visualized in green. The white arrow highlights the biases of the MU and MVU-fix to the curvilinear ribs in iteration 1.}
\label{uncertaintyMaps}
\end{figure}
Fig.~\ref{uncertaintyMaps} shows different uncertainty maps and their corresponding queried regions in iteration 1 and iteration 8 during the training process.
For iteration 1 (i.e., the early training stage), MU and MVU-fix show noticeably high uncertainties for curvilinear background structures and thus cause biased querying operations for them, especially for ribs as highlighted by white arrows. In contrast, MVU-ada is similar to VU in iteration 1 and performs a direct querying for target coronary arteries that occupy only a small set of pixels in XA. Such direct querying for sparse vessels is expected to be more essential in the early training stage for rapidly improving model discrimination capability.
For iteration 8 (i.e., the later training stage), VU generates the uncertainty map that is independent of the callbacks from model updating. It queries even highly confident regions with respect to a relatively mature segmentation model, challenging the model fine-tuning towards a higher convergent accuracy. Owing to the proposed dynamic tradeoff, MVU-ada consistently achieves the appropriate uncertainty maps and thus enables the efficient manual intervention throughout the entire training process. Its corresponding querying operations also focus on challenging regions such as bifurcations, thin branches and terminal vessels with attenuated contrast, effectively correcting the systematic biases of noisy pseudo labels.

\begin{figure*}[t]
\centerline{\includegraphics[width=7in]{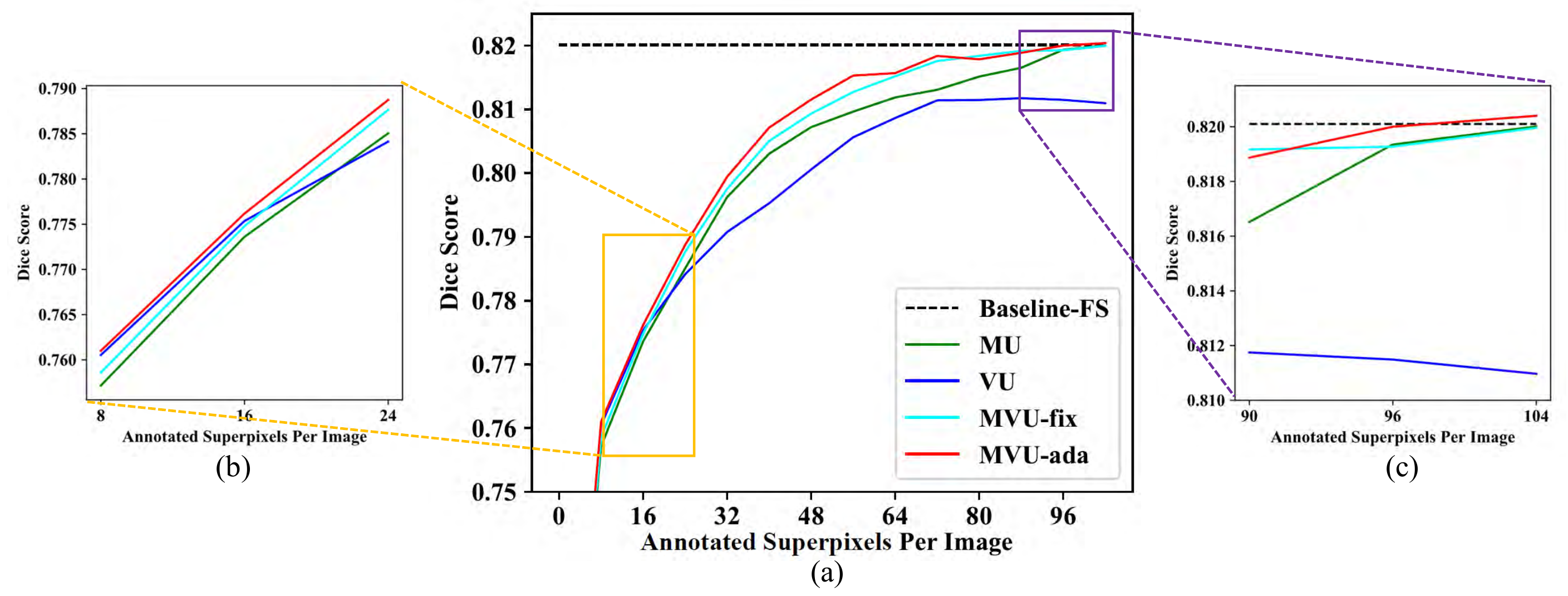}}
\caption{Evolution of dice score with respect to the annotated superpixels that are queried based on different uncertainty estimations and involved incrementally in model training. Subfigure (a) shows the whole training process while subfigure (b) and (c) focus on the early stage and the final model convergence, as bounded by the yellow and purple box, respectively.}
\label{diceScore}
\end{figure*}

To investigate interaction efficiency of suggestive annotation with different uncertainties, as shown in Fig.~\ref{diceScore}, we evaluate the segmentation performance on test images with respect to the incremental annotated superpixels in each training image.
The segmentation performance is gradually improved with the increase of annotations, providing an incremental training process rather than the typical fully supervised framework without manual intervention. Vesselness uncertainty and model uncertainty show the complementary strengths. Specifically, VU has a more rapid performance improvement than MU in the early training stage, as shown in Fig.~\ref{diceScore}(b), indicating that vesselness uncertainty provides the context-aware cue that is more efficient to guide model updating in an early stage. In contrast, MU converges to a higher dice score than VU, as shown in Fig.~\ref{diceScore} (c), owing to the model uncertainty for facilitating model fine-tuning towards a higher convergent accuracy level. MVU-ada and MVU-fix combine both of these uncertainties, exhibiting higher convergent accuracy than VU and faster convergence rate than MU. Among all counterparts, MVU-ada benefits from the dynamic tradeoff between model uncertainty and vesselness uncertainty, leading to the best convergent accuracy and the fastest convergence rate during model training. Its convergent accuracy is even comparable to Baseline-FS that relies on tedious fully manual annotation. Moreover, the fastest convergence rate of MVU-ada not only enables a minimal set of annotations to reach the convergence accuracy (as shown in Fig.~\ref{diceScore}(c)), but also maintains the most efficient manual interaction before model convergence (as shown in Fig.~\ref{diceScore}(a)), i.e., higher dice score with even fewer annotations. It further provides a promising application of this incremental training process in a more cost-effective interactive scenario, where once the desired accuracy is reached even before model convergence, MVU-ada requires the minimal annotation cost and the training process can be early stopped without more annotations further involved.

\begin{figure}[t]
    \centering{\includegraphics[width=3in]{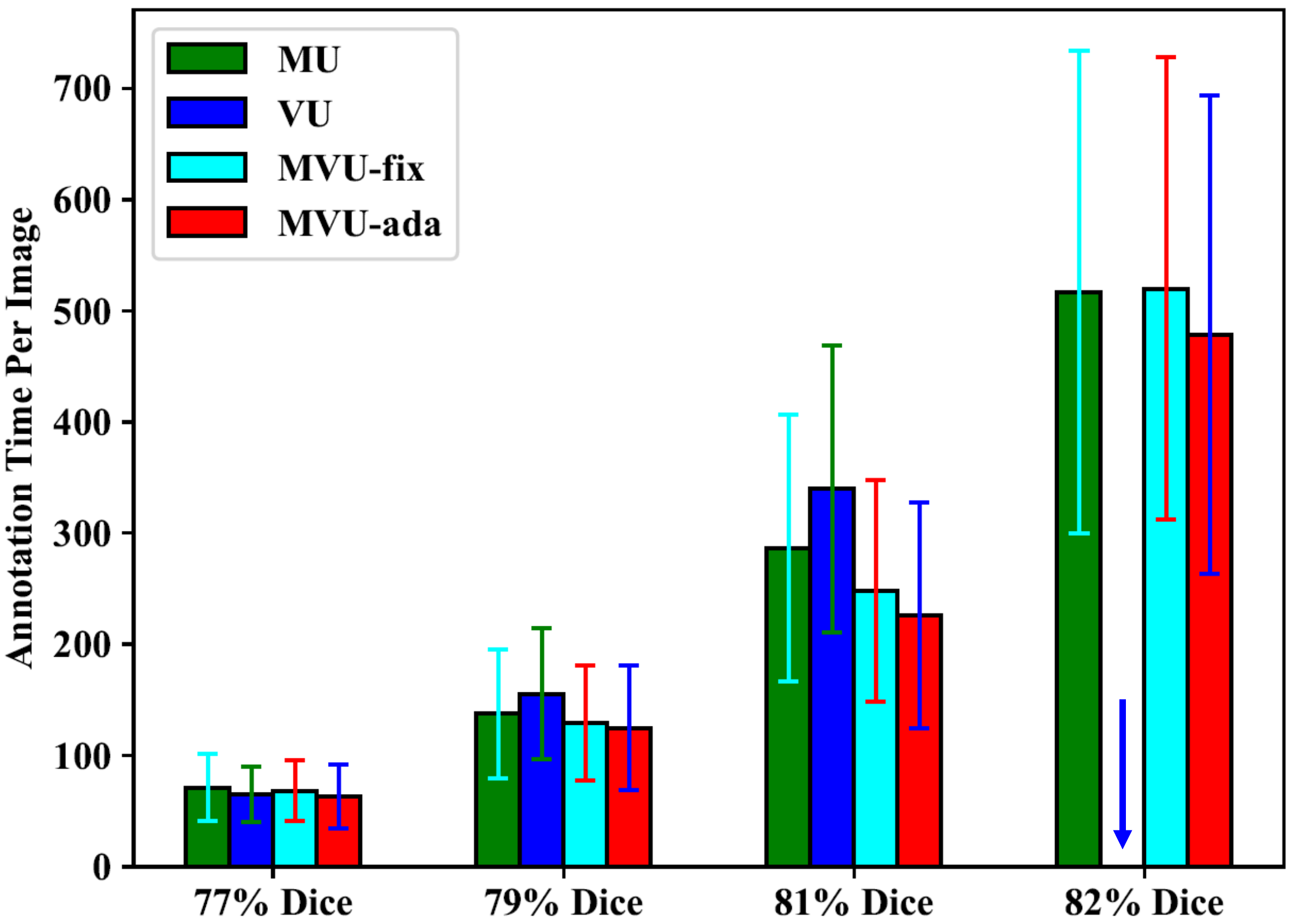}}
    \caption{A comparison of annotation time per image with respect to different desired dice scores. The blue arrow illustrates that VU cannot converge to 82\% dice score.}
    \label{annotationCost}
\end{figure}

\begin{figure*}[t]
\centerline{\includegraphics[width=7in]{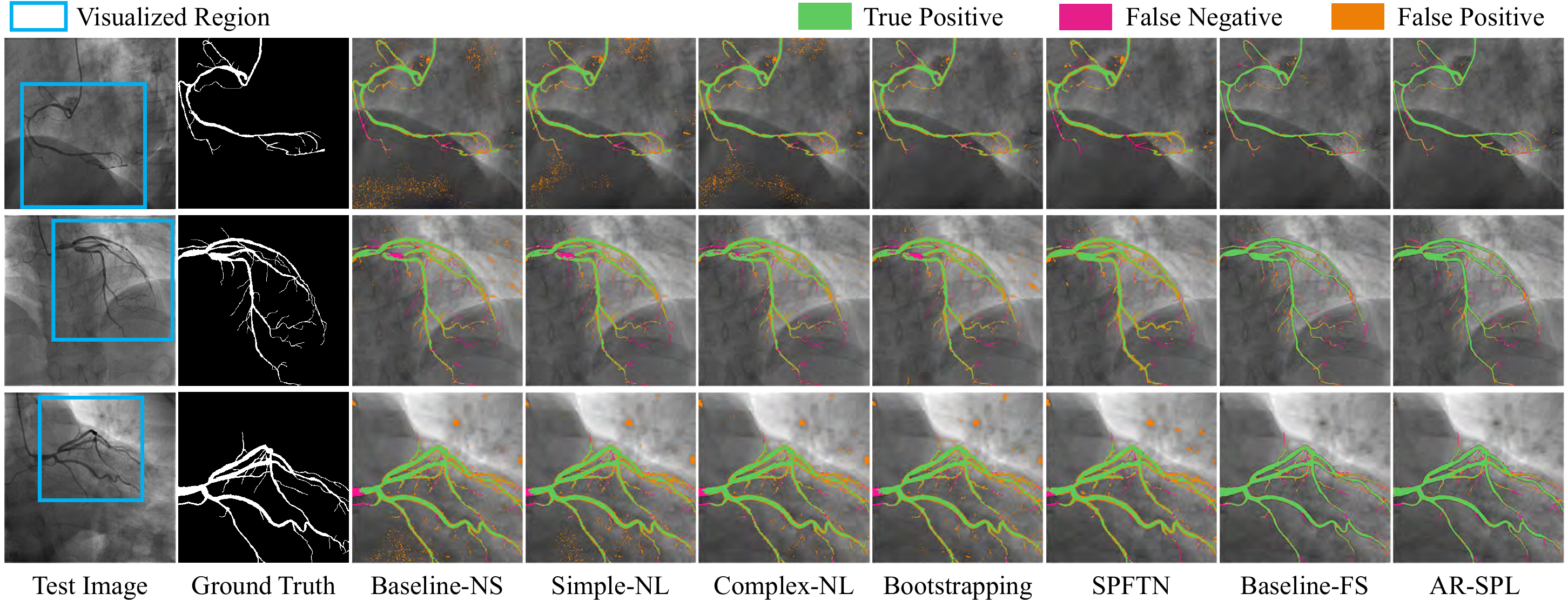}}
\caption{Visual comparison of different weakly supervised learning frameworks and baselines for vessel segmentation in XAs bounded by blue bounding boxes. Except for AR-SPL and Baseline-FS, which respectively utilize sparse and fully manual annotations, the other weakly supervised methods lead to obvious false negatives (highlighted in red) and false positives (highlighted in orange).}
\label{WeaklySupervised}
\end{figure*}

Furthermore, for different uncertainty estimations, Fig.~\ref{annotationCost} provides a detailed comparison for their required annotation times with respect to different accuracy levels. VU costs slightly less annotation time than MU and MVU-fix when model accuracy is fairly low in the early training stage, i.e., 77\% Dice, yet it significantly degenerates when the model accuracy is improved. It even cannot finally converged to 82\% Dice, as highlighted by the blue arrow. Among all counterparts, the proposed MVU-ada consistently requires the least annotation time for different dice scores, indicating its fastest convergence rate, i.e., the most efficient manual interaction, during model training. Especially for the convergent result 82\% Dice, MVU-
ada shows noticeably less annotation time than MU and MVU-fix owing to the proposed dynamic tradeoff between model uncertainty and vesselness uncertainty.

\subsection{Comparison with Other Weakly Supervised Learning Frameworks}\label{cap_weaklySupervised}
We further validate the significance of manual refinement when learning from noisy pseudo labels. The proposed AR-SPL is compared with noisy supervision-based training framework (Baseline-NS), fully supervision-based training framework (Baseline-FS), and the other state-of-the-art weakly supervised learning frameworks that deal with noisy labels without the sparse annotations-based manual refinement proposed in this work:
\begin{itemize}
    \setlength{\itemsep}{0pt}
    \setlength{\parsep}{0pt}
    \setlength{\parskip}{2pt}
    \item Simple noise layer (\textbf{Simple-NL}) \cite{simplenl}, which reduces label noise via an explicit noise model, i.e., a linear layer on the top of the softmax output.
    \item Complex noise layer (\textbf{Complex-NL}) \cite{complexnl}, which improves the explicit noise model in Simple-NL by an additional dependence on feature vectors.
    \item \textbf{Bootstrapping} \cite{bootstrapping}, which augments noisy labels with a notion of perceptual consistency by a convex combination with the current prediction of the model.
    \item Self-paced fine-tuning network (\textbf{SPFTN}) \cite{SPFTN}, which excludes potential label noise by a predefined diversity-based self-paced regularizer.
\end{itemize}

Fig.~\ref{WeaklySupervised} shows the segmentation results of CNNs trained in the different frameworks. Compared with Baseline-NS that applies no noise-robust training strategy, Simple-NL and Complex-NL achieve fewer false positives for semi-transparent background structures owing to their adopted noise models. Bootstrapping and SPFTN further reduce false positives while leaving intractable false negatives for coronary arteries.
When introducing sparse annotations in training images, AR-SPL prominently improves the segmentation performance and approaches Baseline-FS. It enables the fine extraction of the complete coronary trees without increasing false positives in the background.

\begin{table}[t]
    \centering
    \caption{The performances of different weakly supervised methods and baselines. The best performance is highlighted in bold, and its comparable performance is denoted by superscript $^{\textbf{c}}$ based on a two-sided Wilcoxon signed-rank test ($p$-value$>$0.05).}
    \scalebox{1}{\begin{tabular}{llll}
    \hline
    \footnotesize Method & \footnotesize Recall ({\%}) & \footnotesize Precision ({\%}) & \footnotesize Dice ({\%})\\
    \hline
    \footnotesize Baseline-NS & \footnotesize 83.56$\pm$7.28 & \footnotesize 43.60$\pm$10.86 & \footnotesize 56.19$\pm$9.60\\
    \footnotesize Simple-NL & \footnotesize 83.20$\pm$7.42 & \footnotesize 49.97$\pm$12.13 & \footnotesize 61.04$\pm$11.01\\
    \footnotesize Complex-NL & \footnotesize 83.37$\pm$7.75 & \footnotesize 50.84$\pm$13.48 & \footnotesize 61.40$\pm$12.16\\
    \footnotesize Bootstrapping & \footnotesize 83.21$\pm$6.81 & \footnotesize 53.81$\pm$9.82 & \footnotesize 64.66$\pm$7.90\\
    \footnotesize SPFTN & \footnotesize \textbf{85.80$\pm$6.08} & \footnotesize 51.51$\pm$7.34 & \footnotesize 64.10$\pm$6.57\\
    \footnotesize Baseline-FS & \footnotesize 81.44$\pm$4.26 & \footnotesize 82.58$\pm$5.35$^{\textbf{c}}$ & \footnotesize 82.01$\pm$4.21$^{\textbf{c}}$\\
    \footnotesize AR-SPL & \footnotesize 81.64$\pm$4.11 & \footnotesize \textbf{82.69$\pm$5.31} & \footnotesize \textbf{82.09$\pm$4.08}\\
    \hline
    \end{tabular}}
    \label{weaklySupervisedEvaluation}
\end{table}

A quantitative comparison is presented in Table~\ref{weaklySupervisedEvaluation}. The very limited advantage of classical weakly supervised frameworks can be observed for their small superiorities over Baseline-NS. Despite a slightly higher recall, they all suffer from significantly lower precision and dice score than Baseline-FS by a large margin. This performance gap can be bridged when sparse annotation-based manual refinement is involved in the way proposed by AR-SPL, leading to a comparable performance to Baseline-FS without statistically significant difference.

\section{Discussion}\label{discussion}

Precisely annotating coronary arteries in XAs is extremely labor-intensive and time-consuming to train CNNs for vessel segmentation. Noisy pseudo labels generated from vessel enhancement provide an imprecise alternative that is free of manual interaction, yet challenging the accurate segmentation performance at test time. A practical problem is raised: how to learn from these noisy pseudo labels safely without performance deterioration.
Our work offers the first attempt to solve this problem from a novel weakly supervised perspective using ``inaccurate" supervision (i.e., noisy pseudo labels generated from vessel enhancement) together with ``incomplete" supervision (i.e., sparse manual annotations suggested by model-vesselness uncertainty). These two types of weak supervision are leveraged compactly via the proposed AR-SPL, where noisy pseudo labels are first obtained by vessel enhancement, and then refined simultaneously based on self-paced learning and sparse annotations, in order to train an accurate segmentation model with minimal manual intervention.
The experimental results indicate that, despite the very limited annotation cost, our AR-SPL accomplishes the precise vessel extraction and the effective suppression of background disturbance, which is even comparable to the fully supervised learning.

Under a standard weakly supervised learning paradigm, some works use only pseudo labels, and then refine label noise without human interaction by leveraging noise-robust prior knowledge, such as the learning difficulty ~\cite{prior-knowledge,SPFTN} and perceptual consistency ~\cite{bootstrapping}.
However, the experimental results in Section~\ref{cap_weaklySupervised} demonstrate that using such prior knowledge alone causes noticeable performance deterioration at test time. The reasons are two-fold. On one hand, noise-robust prior knowledge would lead to a biased training process and cause overfitting of CNNs on pixels selected by the predefined criterion. On the other hand, pseudo labels generated by vessel enhancement contain inevitable systematic errors, which imply the intractable knowledge defect \cite{idnknow} and hamper the accurate model training. Our proposed AR-SPL provides a simple but powerful solution by using sparse annotations. Sparse manual annotations act as an additional cue to progressively enrich the training variety and compensate knowledge defect in noisy pseudo labels, effectively improving the segmentation performance to a level comparable to fully supervised learning.


In addition, some other similar works \cite{ASPL,repu} follow the self-paced learning under an active learning paradigm, which accepts only sparse manual annotations instead of exploiting noisy pseudo labels. These annotations are suggested by the classical model uncertainty without a context-awareness of vascular geometric features in XAs. Differently from this active learning perspective, the proposed AR-SPL additionally leverages pseudo labels generated from vessel enhancement, and guides sparse manual annotations with a well-designed model-vesselness uncertainty. Our framework offers several advantages. First, pseudo labels help avoid the challenge of collecting initial annotations that would be labor-intensive and time-consuming. They provide a noisy set of initial annotations and thus can be used as a warm-start in the active learning paradigm, which is free of manual intervention and empirical settings. Second, our AR-SPL framework can reduce the requirement for human-provided ground truth by exploiting prior knowledge in pseudo labels such as the intensity distribution and scale information. This knowledge can be obtained by vessel enhancement without labor cost, leading to a beneficial feature representation for coronary arteries in CNNs. Third, the adopted model-vesselness uncertainty takes a customized consideration of vessel structures and facilitates a more efficient manual intervention than the classical model uncertainty, as shown in Section \ref{cap_uncertaintyEstimation}.

To ensure the cost-effective manual intervention, we further investigate the opportunities to alleviate the user's annotation burden using a suggestive annotation strategy with the proposed model-vesselness uncertainty. One single uncertainty cannot maintain the best interaction efficiency during model training.
For example, model uncertainty contributes to the cautious model fine-tuning for a higher convergent accuracy, while hampered by a slow convergence rate due to query redundancy, especially for an early training stage. 
In contrast, vesselness uncertainty provides a context-aware cue for model updating, contributing to solving query redundancy and rapidly improving segmentation performance in the early stage. However, it fails to achieve an effective model fine-tuning and thus leads to poor convergent accuracy due to its independence from the online training process.
The proposed model-vesselness uncertainty incorporates these complementary strengths by leveraging a dynamic time-dependent tradeoff. Specifically, we assign a higher weight to vesselness uncertainty in the early stage for rapid convergence, and then increase the weight for model uncertainty in the later training stage for a better convergent result. 
This well-designed combination strategy consistently minimizes the annotation cost across the entire training process, enabling a cost-effective interactive scenario.

The inflow of contrast agent is substantially affected by lumen diameters and topological variations, leading to attenuated contrast and thus large uncertainty regarding bifurcation points and thin vessels. Manual refinement is suggested for these regions as shown in Fig.~\ref{uncertaintyMaps}, indicating a relatively concentrated distribution instead of scattering over the complete coronary tree. Specifically, queried annotations focus more on terminal branches with small lumen diameters and topologically varying bifurcation points. This concentrated distribution potentially simplifies manual interaction and exhibits higher efficiency versus extensive proofreading over the whole image.

Our method focuses on the vessel segmentation task in a specific medical application, i.e., the PCI surgical planning for coronary artery disease. In the experiments, we have successfully tested our method on the clinical X-ray angiograms during PCI procedure. Nonetheless, application of our method to other vessels may need application-specific modifications. For example, for retinal vessel segmentation in color fundus images, the adopted layer separation for pseudo label generation in our method would be highly limited due to the indistinct motion cue of retinal vessels. A topology constraint \cite{Shi2018Vessel} concerning branch length and lumen diameter would be promising to improve pseudo label generation for retinal vessels. Moreover, the suggestive annotation with model-vesselness uncertainty requires additional designs since vascular features vary widely for different organs and imaging protocols. It would be of interest in the further to investigate the feasibility of adopting our method to vessels from other organs and modalities.

In the experiments, we have validated our proposed method on a clinical X-ray angiogram dataset, where the size is relatively small as compared with many large-scale datasets of natural images such as PASCAL VOC~\cite{everingham2010pascal}, COCO \cite{lin2014microsoft} and ImageNet \cite{deng2009imagenet}. For the segmentation of coronary arteries in XA, it is especially difficult to collect a very large dataset, since pixel-wise manual annotations are highly labor-intensive and require special expertise regarding the thin tubular lumen and complex topology of vessel tree. Therefore, our relatively small dataset is feasible to investigate annotation efficiency, which fits well with our motivation of reducing annotation cost in clinical practice. Nonetheless, it may be unclear whether the generalization capability of the proposed AR-SPL would be affected by the size of dataset. The further work would validate our method on a larger dataset, including more patients, more angiographic viewing angles and different concentrations of injected contrast agent.
In addition, despite the incremental fine-tuning, the proposed AR-SPL is still limited by a considerable time interval between two manual interventions since the updating of model parameters in Eq.~\ref{updatingModel} requires relatively long time. In the further, it also would be of interest to develop a continuous workflow for suggestive annotation by scholastic gradient partial descent (SPADE) \cite{MentorNet} and online learning scheme \cite{onlineDL}.

\section{Conclusions}
In this work, we propose a novel weakly supervised training framework for vessel segmentation in XAs in order to safely and efficiently learn from noisy pseudo labels generated by vessel enhancement. Towards the safety of learning without performance deterioration, label noise is handled effectively by the proposed AR-SPL, where sparse manual annotations provide online guidance for the naive self-paced learning. Furthermore, towards intervention efficiency with minimal annotation cost, we propose a model-vesselness uncertainty with a dynamic tradeoff for suggestive annotation, based on the geometric vesselness and the CNN trained on the fly. Experiments show that compared with fully supervised learning, the proposed AR-SPL achieves very similar segmentation accuracy, and only costs 24.82\% of the annotation time to label 3.46\% of the image regions. Such highly reduced annotation cost reliably alleviates the burden on the annotator and brings about potential advantages in clinical applications for PCI surgical planning, such as a segmentation-based stenosis detection and reconstruction.

\section{Acknowledgment}
This research is partially supported by the National Key research and development program(No. 2016YFC0106200),  Beijing Municipal Natural Science Foundation (No. L192006) , and the funding from Institute of Medical Robotics of Shanghai Jiao Tong University as well as the 863 national research fund (No. 2015AA043203).

\bibliography{library}
\end{document}